\documentclass[11pt, a4paper]{custom_class}

\usepackage{custom_style}
\usepackage[utf8]{inputenc} 
\usepackage[T1]{fontenc}    
\usepackage{hyperref}       
\usepackage{url}            
\usepackage{booktabs}       
\usepackage{amsfonts}       
\usepackage{nicefrac}       
\usepackage{microtype}      
\usepackage{cleveref}       
\usepackage{lipsum}         
\usepackage{graphicx}
\usepackage{natbib}
\usepackage{doi}
\usepackage{glossaries}
\title{A Philosophical Introduction to Language Models \\[1ex] \subtitlefont\sc Part I: Continuity With Classic Debates}

\author{
Raphaël Millière \\
Department of Philosophy \\
Macquarie University \\
\texttt{raphael.milliere@mq.edu.eu} \\
\And
Cameron Buckner \\
Department of Philosophy \\
University of Houston \\
\texttt{cjbuckner@uh.edu} \\
}

\renewcommand{\headeright}{Part I}
\renewcommand{\undertitle}{}
\renewcommand{\shorttitle}{A Philosophical Introduction to Language Models}

\definecolor{glossarycolor}{RGB}{107, 0, 0}

\date{} 

\makeglossaries

\newglossaryentry{blockhead}{
  name={Blockhead},
  description={A philosophical thought experiment introduced by \cite{blockPsychologismBehaviorism1981}, illustrating a hypothetical system that mimics human-like responses without genuine understanding or intelligence. Blockhead's responses are preprogrammed, allowing it to answer any conceivable question based on retrieval from an extensive database, akin to a hash table lookup. This system challenges traditional notions of intelligence by demonstrating behaviorally indistinguishable from a human's, yet lacking the internal cognitive processes typically associated with intelligence. Blockhead serves as a critical example in discussions about the nature of artificial intelligence, emphasizing the distinction between mere behavioral mimicry and the presence of complex, internal information processing mechanisms as a hallmark of true intelligence}
}

\newglossaryentry{generalization}{
  name={generalization},
  description={The ability of a neural network model to perform accurately on new, unseen data that is similar but not identical to the data it was trained on. This concept is central to evaluating the effectiveness of a model, as it indicates the extent to which the learned patterns and knowledge can be applied beyond the specific examples in the training dataset. A model that generalizes well maintains high performance when faced with new and varied inputs, demonstrating its adaptability and robustness across a broad range of scenarios}
}

\newglossaryentry{logit}{
  name={logit},
  description={In the context of Transformer-based LLMs, a logit is the raw output of the model's final layer before it undergoes a softmax transformation to become a probability distribution. Each logit corresponds to a potential output token (e.g., a word or subword unit), and its value indicates the model's preliminary assessment of how likely that token is to be the next element in the sequence, given the input. The softmax function then converts these logits into a probability distribution, from which the model selects the most likely next token during text generation}
}

\newglossaryentry{ooddata}
{
    name={out-of-distribution (OOD) data},
    description={In machine learning, OOD data refers to input data that significantly differs from the data the model was trained on. This type of data falls outside the distribution of the training dataset, presenting patterns, features, or characteristics that the model has not encountered during its training phase. OOD data is a critical concept because it challenges the model's ability to generalize and maintain accuracy. Handling OOD data effectively is important for robustness and reliability, especially in real-world applications where the model is likely to encounter a wide variety of inputs}
}

\newglossaryentry{selfattention}
{
    name={self-attention},
    description={A mechanism within \Gls{transformer}-based neural networks that enables them to weigh and integrate information from different positions within the input sequence. In the context of LLMs, self-attention allows each token in a sentence to be processed in relation to every other token, facilitating the understanding of context and relationships within the text. This process involves calculating attention scores that reflect the relevance of each part of the input to every other part, thereby enhancing the model's ability to capture dependencies, regardless of their distance in the sequence. This feature is key to LLMs' ability to handle long-range dependencies and complex linguistic structures effectively}
}

\newglossaryentry{tokenization}
{
    name={tokenization},
    description={The process of breaking down text into smaller units, called tokens. These tokens can be words, subwords, characters, or other meaningful elements, depending on the granularity of the tokenization algorithm. The purpose of tokenization is to transform the raw text into a format that can be easily processed and understood by a language model. This step is crucial for preparing input data, as it directly affects the model's ability to analyze and generate language. Tokenization plays a fundamental role in determining the level of detail and complexity a model can capture from the text, but can also have a downstream impact on the model's performance with certain tasks such as arithmetic}
}

\newglossaryentry{traintestsplit}
{
    name={train-test split},
    description={In machine learning, the train-test split  is a method used to evaluate the performance of a model. It involves dividing the available data into two distinct sets: a training set and a test set. The training set is used to train the model, allowing it to learn and adapt to patterns within the data. The test set, which consists of data not seen by the model during its training, is used to assess the model's performance and \gls{generalization} capabilities. This split is crucial for providing an unbiased evaluation of the model, as it demonstrates how the model is likely to perform on new, unseen data}
}

\newglossaryentry{transformer}
{
    name={Transformer},
    description={A type of neural network architecture introduced by \cite{vaswaniAttentionAllYou2017}, predominantly used for processing sequential data such as text. It is characterized by its reliance on \gls{selfattention} mechanisms, which enable it to weigh the importance of different parts of the input data. Unlike earlier architectures, Transformers do not require sequential data to be processed in order, allowing for more parallel processing and efficiency in handling long-range dependencies in data. This architecture forms the basis of most LLMs, known for its effectiveness in capturing complex linguistic patterns and relationships}
}

\newglossaryentry{vector}
{
    name={vector},
    description={Mathematically, a \gls{vector} is an ordered array of numbers, which can represent points in a multidimensional space. In the context of LLMs, \glspl{vector} are used to represent tokens, where each token can map onto a word or part of a word depending on the \gls{tokenization} scheme. These \glspl{vector}, known as embeddings, encode the linguistic features and relationships of the tokens in a high-dimensional space. By converting tokens into \glspl{vector}, LLMs are able to process and generate language based on the semantic and syntactic properties encapsulated in these numerical representations}
}

\begin{document}
\maketitle

\begin{abstract}
Large language models like GPT-4 have achieved remarkable proficiency in a broad spectrum of language-based tasks, some of which are traditionally associated with hallmarks of human intelligence. This has prompted ongoing disagreements about the extent to which we can meaningfully ascribe any kind of linguistic or cognitive competence to language models. Such questions have deep philosophical roots, echoing longstanding debates about the status of artificial neural networks as cognitive models. This article--the first part of two companion papers--serves both as a primer on language models for philosophers, and as an opinionated survey of their significance in relation to classic debates in the philosophy cognitive science, artificial intelligence, and linguistics. We cover topics such as compositionality, language acquisition, semantic competence, grounding, world models, and the transmission of cultural knowledge. We argue that the success of language models challenges several long-held assumptions about artificial neural networks. However, we also highlight the need for further empirical investigation to better understand their internal mechanisms. This sets the stage for the companion paper (Part II), which turns to novel empirical methods for probing the inner workings of language models, and new philosophical questions prompted by their latest developments.
\end{abstract}

\section{Introduction} \label{sec:intro}

Deep learning has catalyzed a significant shift in artificial intelligence over the past decade, leading up to the development of Large Language Models (LLMs). The reported achievements of LLMs, often heralded for their ability to perform a wide array of language-based tasks with unprecedented proficiency, have captured the attention of both the academic community and the public at large. State-of-the-art LLMs like GPT-4 are even claimed to exhibit ``sparks of general intelligence'' \citep{bubeckSparksArtificialGeneral2023}. They can produce essays and dialogue responses that often surpass the quality of an average undergraduate student's work \citep{herboldLargescaleComparisonHumanwritten2023}; they achieve better scores than most humans on a variety of AP tests for college credit and rank in the 80-99th percentile on graduate admissions tests like the GRE or LSAT \citep{openaiGPT4TechnicalReport2023}; their programming proficiency ``favorably compares to the average software engineer's ability'' \citep{bubeckSparksArtificialGeneral2023,savelkaThrilledYourProgress2023}; they can solve many difficult mathematical problems \citep{zhouSolvingChallengingMath2023a}--even phrasing their solution in the form of a Shakespearean sonnet, if prompted to do so. LLMs also form the backbone of multimodal systems that can answer advanced questions about visual inputs \citep{openaiGPT4VIsionSystem2023} or generate images that satisfy complex compositional relations based on linguistic descriptions \citep{betker2023improving}.\footnote{GPT-4V \citep{openaiGPT4VIsionSystem2023} is a single multimodal model that can take both text and images as input; by contrast, DALL-E 3 \citep{betker2023improving} is a distinct text-to-image model that can be seamlessly prompted by GPT-4--an example of model ensembling using natural language as a universal interface \citep{zengSocraticModelsComposing2022}. While officially available information about GPT-4, GPT-4V and DALL-E 3 is scarce, they are widely believed to use a \Gls{transformer} architecture as backbone to encode linguistic information, like similar multimodal models and virtually all LLMs \citep{brownLanguageModelsAre2020,touvronLlamaOpenFoundation2023,rameshHierarchicalTextConditionalImage2022,alayracFlamingoVisualLanguage2022}.} While the released version of GPT-4 was intentionally hobbled to be unable to perfectly imitate humans--to mitigate plagiarism, deceit, and unsafe behavior--it nevertheless still managed to produce responses that were indistinguishable from those written by humans at least 30\% of the time when assessed on a two-person version of the Turing test for intelligence \citep{jonesDoesGPT4Pass2023}. This rate exceeds the threshold established by Turing himself for the test: that computer programs in the 21st century should imitate humans so convincingly that an average interrogator would have less than a 70\% chance of identifying them as non-human after five minutes of questioning \citep{turingComputingMachineryIntelligence1950}.

To philosophers who have been thinking about artificial intelligence for many years, GPT-4 can seem like a thought experiment come to life--albeit one that calls into question the link between intelligence and behavior. As early as 1981, Ned Block imagined a hypothetical system--today commonly called ``\Gls{blockhead}''--that exhibited behaviors indistinguishable from an adult human's, yet was not considered intelligent.\footnote{Key technical terms in this paper are highlighted in \textcolor{glossarycolor}{red} and defined in the \hyperref[glossary]{glossary}. In the electronic version, these terms are interactive and link directly to their respective glossary entries.} Block's challenge focused on the way in which the system produced its responses to inputs. In particular, \Gls{blockhead}'s responses were imagined to have been explicitly preprogrammed by a ``very large and clever team [of human researchers] working for a very long time, with a very large grant and a lot of mechanical help,'' to devise optimal answers to any potential question the judge might ask \citep[][p. 20]{blockPsychologismBehaviorism1981}. In other words, \Gls{blockhead} answers questions not by understanding the inputs and processing them flexibly and efficiently, but rather simply retrieving and regurgitating the answers from its gargantuan memory, like a lookup operation in a hash table. The consensus among philosophers is that such a system would not qualify as intelligent. In fact, many classes in the philosophy of artificial intelligence begin with the position Block and others called ``psychologism:'' intelligence does not merely depend on the observable behavioral dispositions of a system, but also on the nature and complexity of internal information processing mechanisms that drive these behavioral dispositions.

In fact, many of GPT-4's feats may be produced by a similarly inefficient and inflexible memory retrieval operation. GPT-4's training set likely encompasses trillions of tokens in millions of textual documents, a significant subset of the whole internet.\footnote{While details about the training data of GPT-4 are not publicly available, we can turn to other LLMs for clues. For example, PaLM 2 has 340 billion parameters and was trained on 3.6 trillion tokens \citep{anilPaLMTechnicalReport2023}, while the largest version of Llama 2 has 70 billion parameters and was trained on 2 trillion tokens \citep{touvronLlamaOpenFoundation2023}. GPT-4 is rumored to have well over a trillion parameters \citep{karhadeGPT4ModelsOne2023}.} Their training sets include dialogues generated by hundreds of millions of individual humans and hundreds of thousands of academic publications covering potential question-answer pairs. Empirical studies have discovered that the many-layered architecture of DNNs grants them an astounding capacity to memorize their training data, which can allow them to retrieve the right answers to millions of randomly-labeled data points in artificially-constructed datasets where we know a priori there are no abstract principles governing the correct answers \citep{zhangUnderstandingDeepLearning2021}. This suggests that GPT-4's responses could be generated by approximately--and, in some cases, exactly--reproducing samples from its training data.\footnote{This concern is highlighted by lawsuits against OpenAI, notably from the New York Times \citep{grynbaumTimesSuesOpenAI2023}. These cases document instances where LLMs like GPT-4 have been shown to reproduce substantial portions of copyrighted text verbatim, raising questions about the originality of their outputs.} If this were all they could do, LLMs like GPT-4 would simply be \Glspl{blockhead} come to life. Compare this to a human student who had found a test's answer key on the Internet and reproduced its answers without any deeper understanding; such regurgitation would not be good evidence that the student was intelligent. For these reasons, ``data contamination''--when the training set contains the very question on which the LLM's abilities are assessed--is considered a serious concern in any report of an LLM's performance, and many think it must be ruled out by default when comparing human and LLM performance \citep{aiyappaCanWeTrust2023}. Moreover, GPT-4's pre-training and fine-tuning requires an investment in computation on a scale available only to well-funded corporations and national governments--a process which begins to look quite inefficient when compared to the data and energy consumed by the squishy, 20-watt engine between our ears before it generates similarly sophisticated output.

In this opinionated review paper, we argue that LLMs are more than mere \Glspl{blockhead}; but this skeptical interpretation of LLMs serves as a useful foil to develop a subtler view. While LLMs \textit{can} simply regurgitate large sections of their prompt or training sets, they are also capable of flexibly blending patterns from their training data to produce genuinely novel outputs. Many empiricist philosophers have defended the idea that sufficiently flexible copying of abstract patterns from previous experience could form the basis of not only intelligence, but full-blown creativity and rational decision-making \citep{baierHumeReflectiveWomen2002, humeTreatiseHumanNature1978,bucknerDeepLearningRational2023}; and more scientific research has emphasized that the kind of flexible \gls{generalization} that can be achieved by interpolating \glspl{vector} in the semantic spaces acquired by these models may explain why these systems often appear more efficient, resilient, and capable than systems based on rules and symbols \citep{smolenskyProperTreatmentConnectionism1988, smolenskyNeurocompositionalComputingCentral2022}. A useful framework for exploring the philosophical significance of such LLMs, then, might be to treat the worry that they are merely unintelligent, inefficient \Glspl{blockhead} as a null hypothesis, and survey the empirical evidence that can be mustered to refute it.\footnote{Such a method of taking a deflationary explanation for data as a null hypothesis and attempting to refute it with empirical evidence has been a mainstay of comparative psychology for more than a century, in the form of Morgan's Canon \citep{bucknerUnderstandingAssociativeCognitive2017,soberMorganCanon1998}. As DNN-based systems approach the complexity of an animal brain, it may be useful to take lessons from comparative psychology in arbitrating fair comparisons to human intelligence \citep{bucknerBlackBoxesUnflattering2021}. In comparative psychology, standard deflationary explanations for data include reflexes, innate-releasing mechanisms, and simple operant conditioning. Here, we suggest that simple deflationary explanations for an AI-inspired version of Morgan's Canon include \Gls{blockhead}-style memory lookup.}

We adopt that approach here, and use it to provide a brief introduction to the architecture, achievements, and philosophical questions surrounding state-of-the-art LLMs such as GPT-4. There has, in our opinion, never been a more important time for philosophers from a variety of backgrounds--but especially philosophy of mind, philosophy of language, epistemology, and philosophy of science--to engage with foundational questions about artificial intelligence. Here, we aim to provide a wide range of those philosophers (and philosophically-inclined researchers from other disciplines) with an opinionated survey that can help them to overcome the barriers imposed by the  technical complexity of these systems and the ludicrous pace of recent research achievements.

\section{A primer on LLMs} \label{sec:basic-explanation-of-llms}

\subsection{Historical foundations}

The origins of large language models can be traced back to the inception of AI research. The early history of natural language processing (NLP) was marked by a schism between two competing paradigms: the symbolic and the stochastic approaches. A major influence on the symbolic paradigm in NLP was Noam Chomsky's transformational-generative grammar \citep{chomskySyntacticStructures1957}, which posited that the syntax of natural languages could be captured by a set of formal rules that generated well-formed sentences. Chomsky's work laid the foundation for the development of rule-based syntactic parsers, which leverage linguistic theory to decompose sentences into their constituent parts. Early conversational NLP systems, such as Winograd's SHRDLU \citep{winogradProceduresRepresentationData1971}, required syntactic parsers with a complex set of ad hoc rules to process user input.

In parallel, the stochastic paradigm was pioneered by researchers such as mathematician Warren Weaver, who was influenced by Claude Shannon's information theory. In a memorandum written in 1949, Weaver proposed the use of computers for machine translation employing statistical techniques \citep{weaverTranslation1955}. This work paved the way for the development of statistical language models, such as n-gram models, which estimate the likelihood of word sequences based on observed frequencies of word combinations in a corpus \citep{jelinekStatisticalMethodsSpeech1998}. Initially, however, the stochastic paradigm was lagging behind symbolic approaches to NLP, showing only modest success in toy models with limited applications.

Another important theoretical stepping stone on the road to modern language models is the so-called distributional hypothesis, first proposed by the linguist Zellig Harris in the 1950s \citep{harrisDistributionalStructure1954}. This idea was grounded in the structuralist view of language, which posits that linguistic units acquire meaning through their patterns of co-occurrence with other units in the system. Harris specifically suggested that the meaning of a word could be inferred by examining its distributional properties, or the contexts in which it occurs. \cite{firthSynopsisLinguisticTheory1957} aptly summarized this hypothesis with the slogan ``You shall know a word by the company it keeps,'' acknowledging the influence of \cite{wittgensteinPhilosophicalInvestigations1953}'s conception of meaning-as-use to highlight the importance of context in understanding linguistic meaning.

\begin{figure}[h] 
    \centering 
    \includegraphics[width=1\linewidth]{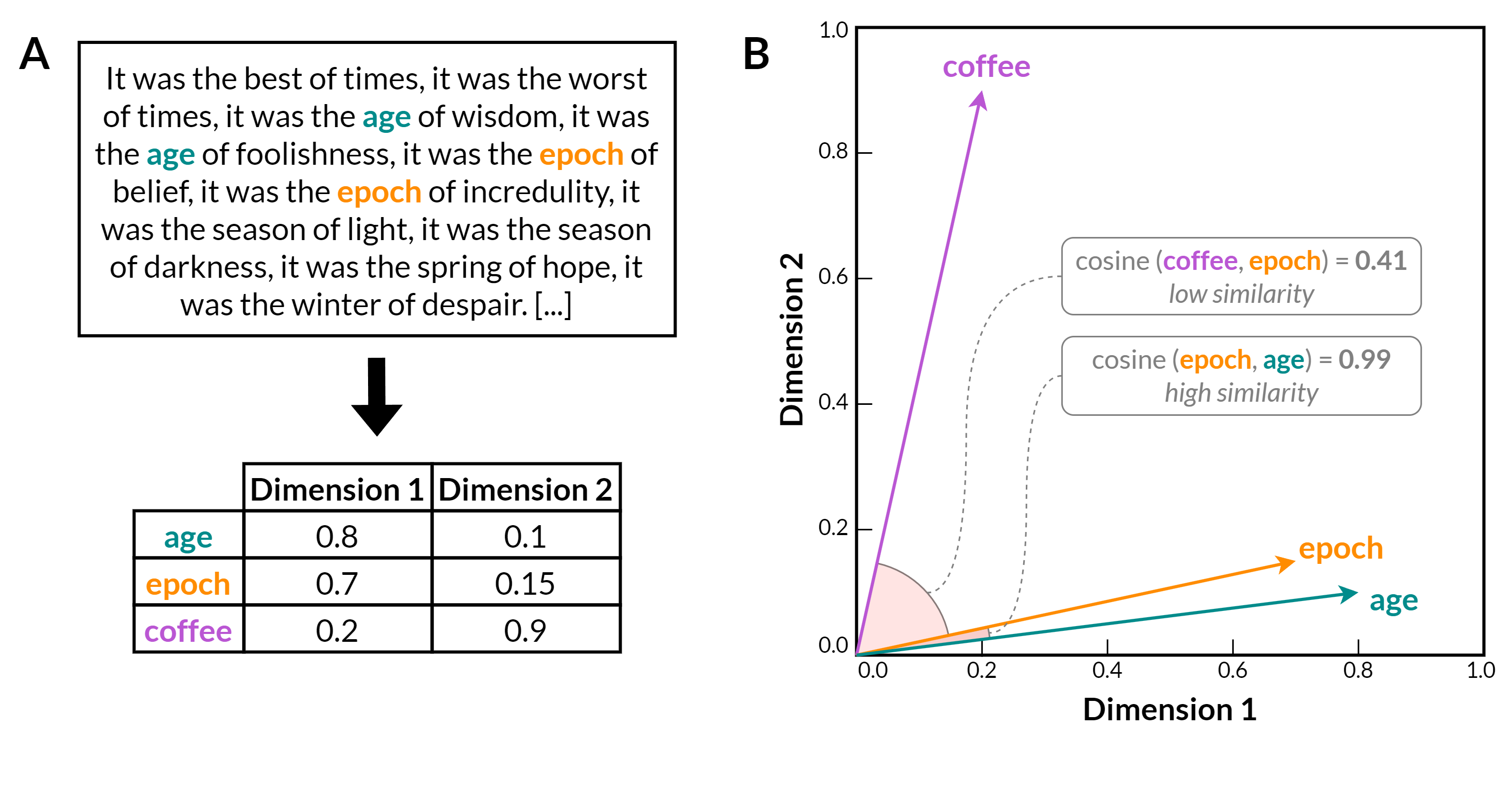} 
    \caption{\textbf{An illustration of word embeddings in a multidimensional vector space.} \textbf{A.} A word embedding model trained on a natural language corpus learns to encode words into numerical \textit{\glspl{vector}} (or \textit{embeddings}) in a multidimensional space (simplified to two dimensions for visual clarity). Over the course of training, vectors for contextually related words (such as `age' and `epoch') become more similar, while vectors for contextually unrelated words (such as `age' and `coffee') become less similar. \textbf{B.} Word embeddings in the two-dimensional vector space of a trained model. Words with similar meanings (`age' and `epoch') are positioned closer together, as indicated by their high cosine similarity score, whereas words with dissimilar meanings (`coffee' and `epoch') are further apart, reflected in a lower cosine similarity score. Cosine similarity is a measure used to determine the cosine of the angle between two non-zero vectors, providing an indication of the degree to which they are similar. A cosine similarity score closer to 1 indicates a smaller angle and thus a higher degree of similarity between the vectors. Figure loosely adapted from \citet[][Figure 1]{boledaDistributionalSemanticsLinguistic2020}.}
    \label{fig:vector_spaces} 
\end{figure}

As research on the distributional hypothesis progressed, scholars began exploring the possibility of representing word meanings as \glspl{vector} in a multidimensional space \ref{fig:vector_spaces}. Early empirical work in this area stemmed from psychology and examined the meaning of words along various dimensions, such as valence and potency \citep{osgoodNatureMeasurementMeaning1952}. While this work introduced the idea of representing meaning in a multidimensional \gls{vector} space, it relied on explicit participant ratings about word connotations along different scales (e.g., \emph{good}--\emph{bad}), rather than analyzing the distributional properties of a linguistic corpus. Subsequent research in information retrieval combined vector-based representations with a data-driven approach, developing automated techniques for representing documents and words as \glspl{vector} in high-dimensional \gls{vector} spaces \citep{saltonVectorSpaceModel1975}.

After decades of experimental research, these ideas eventually reached maturity with the development of word embedding models using artificial neural networks \citep{bengioNeuralProbabilisticLanguage2000}. These models are based on the insight that the distributional properties of words can be \textit{learned} by training a neural network to predict a word's context given the word itself, or vice versa. Unlike previous statistical methods such as n-gram models, word embedding models encode words into dense, low-dimensional \gls{vector} representations (Fig. \ref{fig:vector_spaces}). The resulting \gls{vector} space drastically reduces the dimensionality of linguistic data while preserving information about meaningful linguistic relationships beyond simple co-occurrence statistics. Notably, many semantic and syntactic relationships between words are reflected in linear substructures within the \gls{vector} space of word embedding models. For example, Word2Vec \citep{mikolovEfficientEstimationWord2013} demonstrated that word embeddings can capture both semantic and syntactic regularities, as evidenced by the ability to solve word analogy tasks through simple \gls{vector} arithmetic that reveal the latent linguistic structure encoded in the \gls{vector} space (e.g., $king + woman - man \approx queen$, or $walking + swam - walked \approx swimming$).

The development of word embedding models marked a turning point in the history of NLP, providing a powerful and efficient means of representing linguistic units in a continuous \gls{vector} space based on their statistical distribution in a large corpus. However, these models have several significant limitations. First, they are not capable of capturing polysemy and homonymy, because they assign a single or ``static'' embedding to each word type, which cannot account for changes in meaning based on context; for example, ``bank'' is assigned a unique embedding regardless of whether it refers to the side of a river or the financial institution. Second, they rely on ``shallow'' artificial neural network architectures with a single hidden layer, which limits their ability to model complex relationships between words. Finally, being designed to represent language at the level of individual words, they are not well-suited to model complex linguistic expression, such as phrases, sentences, and paragraphs. While it is possible to represent a sentence as a \gls{vector} by averaging the embeddings of every word in the sentence, this is a very poor way of representing sentence-level meaning, as it loses information about compositional structure reflected in word order. In other words, word embedding models merely treat language as a ``bag of words''; for example, ``a law book'' and ``a book law'' are treated identically as the unordered set \{\texttt{\textquotesingle a\textquotesingle,\textquotesingle book\textquotesingle,\textquotesingle law\textquotesingle}\}.

The shortcomings of shallow word embedding models were addressed with the introduction of ``deep'' language models, going back to recurrent neural networks (RNNs) and their variants, such as long short-term memory (LSTM) \citep{hochreiterLongShortTermMemory1997} and the gated recurrent unit (GRU) \citep{choLearningPhraseRepresentations2014}. These deep neural network architectures incorporate a memory-like mechanism, allowing them to remember and process sequences of inputs over time, rather than individual, isolated words. Despite this advantage over word embedding models, they suffer from their own limitations: they are slow to train and struggle with long sequences of text. These issues were addressed with the introduction of the \Gls{transformer} architecture by \cite{vaswaniAttentionAllYou2017}, which laid the groundwork for modern LLMs. 

\subsection{Transformer-based LLMs}

One of the key advantages of the \Gls{transformer} architecture is that all words in the input sequence are processed in parallel rather than sequentially, by contrast with RNNs, LSTMS and GRUs.\footnote{Note that parallel processing in \Glspl{transformer} is applicable within a predefined maximum sequence length or input window, beyond which the model cannot process without truncating or segmenting the input. This is due to the \gls{selfattention} mechanism's quadratic growth in computational and memory requirements with respect to the sequence length.} Not only does this architectural modification greatly boost training efficiency, it also improves the model's ability to handle long sequences of text, thus increasing the scale and complexity of language tasks that can be performed.

\begin{figure}[h!] 
    \centering 
    \includegraphics[width=0.9\linewidth]{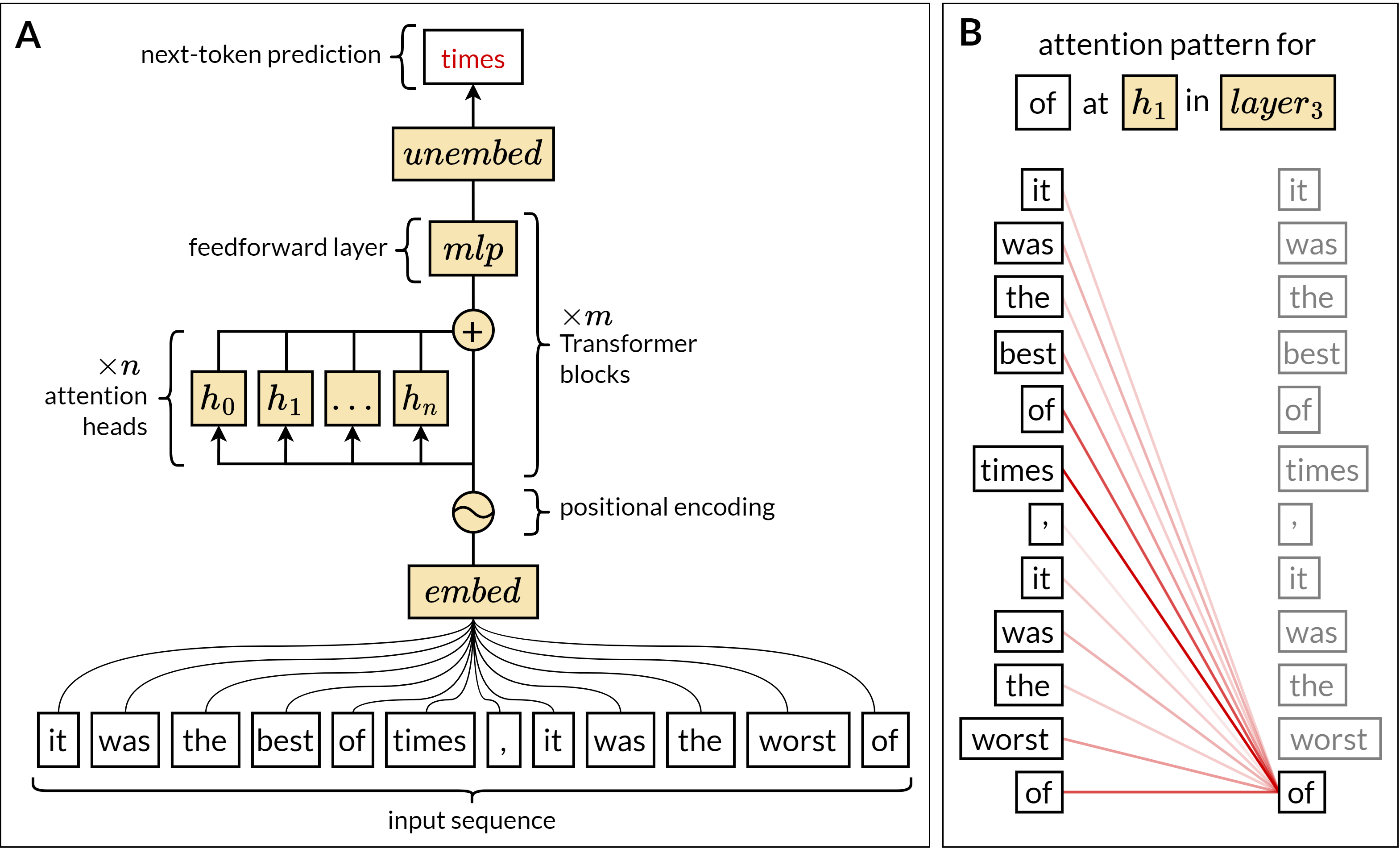} 
    \caption{\textbf{A. The autoregressive Transformer architecture of LLMs.} Tokens from the input sequence are first embedded as \glspl{vector}, which involves converting each token into a high-dimensional space where semantically similar tokens have correspondingly similar \glspl{vector}. Positional encoding adds information about the position of each token in the input sequence to the \glspl{vector}. These enriched \glspl{vector} are then processed through successive \Gls{transformer} blocks. Each block consists of multiple attention heads that process all \glspl{vector} in parallel, and a fully-connected feedforward layer, also known as a multilayer perceptron (MLP) layer. Finally, in the unembedding stage, the \glspl{vector} undergo a linear transformation to project them into a vocabulary-sized space, producing a set of \glspl{logit}. These \glspl{logit} represent the unnormalized scores for each potential next token in the vocabulary. A softmax layer is then applied to convert these \glspl{logit} into a probability distribution over the vocabulary, indicating the comparative likelihood of each token being the next in the sequence. During the training process, the correct next token is known and used for backpropagation, whereas during inference, the model predicts the next token without this information. This process can be repeated iteratively in an autoregressive manner for each token prediction to generate more than one token. \textbf{B. The \gls{selfattention} mechanism visualized.} Each attention head assigns a weight or \textit{attention score} to each token $t_i$ for every token $t_{0-i}$ in the sequence up to and including $t_i$. Here, each red line represents the attention score between `of' and every other token in the input sequence, including itself. In this example, the attention score quantifies the relevance or importance of each token with respect to the token `of', with thicker lines indicating higher scores. This pattern exemplifies how the attention mechanism allows the model to dynamically focus on different parts of the input sequence to derive a contextually nuanced representation of each token. The attention pattern is different for every head, because that each head specializes during training in selectively attending to specific kinds of dependencies between tokens.}
    \label{fig:transformer} 
\end{figure}

At the heart of the \Gls{transformer} model lies a mechanism known as \emph{\gls{selfattention}} (Fig. \ref{fig:transformer}). Simply put, \gls{selfattention} allows the model to weigh the importance of different parts of a sequence when processing each individual word contained in that sequence. For instance, when processing the word ``it'' in a sentence, the \gls{selfattention} mechanism allows the model to determine which previous word(s) in the sentence ``it'' refers to--which can change for different occurrences of ``it'' in different sentences. This mechanism helps LLMs to construct a sophisticated representation of long sequences of text by considering not just individual words, but the interrelationships among all words in the sequence. Beyond the sentence level, it enables LLMs to consider the broader context in which expressions occur, tracing themes, ideas, or characters through paragraphs or whole documents.

It is worth mentioning that \Gls{transformer} models do not operate on words directly, but on linguistic units known as \emph{tokens}. Tokens can map onto whole words, but they can also map onto smaller word pieces. Before each sequence of words is fed to the model, it is chunked into the corresponding tokens -- a process called \emph{\gls{tokenization}}. The goal of \gls{tokenization} is to strike a balance between the total number of unique tokens (the size of the model's ``vocabulary''), which should be kept relatively low for computational efficiency, and the ability to represent as many words from as many languages as possible, including rare and complex words. This is typically accomplished by breaking down words into common sub-word units, which need not carve them at their morphologically meaningful joints. For example, GPT-4's tokenizer maps the word ``metaphysics'' onto three tokens, corresponding to ``met'', ``aph'', and ``ysics'' respectively. Similarly, numbers need not be tokenized into meaningful units for the decimal system; GPT-3 processes ``940'' as a single token, while ``941'' gets processed as two tokens for ``9'' and ``41'' (this quirk of \gls{tokenization} goes a long way towards explaining why LLMs can struggle with multiple-digit arithmetic; see \citet{wallaceNLPModelsKnow2019} and \citet{leeTeachingArithmeticSmall2023}).

The most common variant of Transformer-based models are known as ``autoregressive'' models -- including GPT-3, GPT-4, and ChatGPT. Autoregressive models operate using a learning objective called \textit{next-token prediction}: given a sequence of tokens, they are tasked with predicting which token is statistically most likely to follow. They are trained on a large corpus that includes a diverse range of sources, such as encyclopedias, academic articles, books, websites, and, in more recent iterations of the GPT series, even a substantial amount of computer code. The goal is to provide the model with a rich, diverse dataset that encapsulates the breadth and depth of natural and artificial languages, allowing it to learn from next-token prediction in many different contexts.

At each training step, the model's objective is to predict the next token in a sequence sampled from the corpus, based on the tokens that precede. For instance, given the sequence ``The cat is on the,'' the model might predict ``mat'' as the most likely next token. The model is initialized with random parameters, which means its first predictions are essentially no better than chance. However, with each prediction, the model's parameters are incrementally adjusted to minimize the discrepancy between its prediction and the actual next token in the training data. This process is iterated over billions of training steps, until the model becomes excellent at predicting the next token in any context sampled from the training data. This means that a trained model should be able to write text -- or code -- fluently, picking up on contextual clues provided in the ``prompt'' or input sequence.

While LLMs trained in this way are very good at generating convincing paragraphs, they have no intrinsic preference for truthful, useful, or inoffensive language. Indeed, the task of next-token prediction does not explicitly incorporate common normative goals of human language use. To overcome this limitation, it is possible to refine a model pre-trained with next-token prediction by ``fine-tuning'' it on a different learning objective. One popular fine-tuning technique in recent LLMs such as ChatGPT is called ``reinforcement learning from human feedback,'' or RLHF \citep{christianoDeepReinforcementLearning2017}.

RLHF proceeds in three stages. The initial stage involves collecting a dataset of human comparisons. In this phase, human crowdworkers are asked to review and rank different model responses according to their quality. For instance, the model may generate multiple responses to a particular prompt, and human reviewers are asked to rank these responses based on certain normative criteria such as helpfulness, harmlessness and honesty \citep[the ``three Hs'',][]{askellGeneralLanguageAssistant2021}. This results in a comparison dataset that reflects a preference ranking among possible responses to a set of inputs. In the second stage, this comparison data is used to train a reward model that guides the fine-tuning process of the model. A reward model is a function that assigns a numerical score to a given model output based on its perceived quality. By leveraging the comparison data, developers can train this reward model to estimate the quality of different responses. In the third and final stage, the reward model's outputs are used as feedback signals in a reinforcement learning process, to fine-tune the pre-trained LLM's parameters. In other words, the pre-trained LLM learns to generate responses that are expected to receive higher rankings based on the reward model. This process can be repeated iteratively, such that the model's performance improves with each iteration. However, the effectiveness of this approach heavily relies on the quality of the comparison data and the reward model.

RLHF allows developers to steer model outputs in more specific and controlled directions. For instance, this method can be utilized to reduce harmful and untruthful outputs, to encourage the model to ask clarifying questions when a prompt is ambiguous, or to align the model's responses with specific ethical guidelines or community norms. By combining next-token prediction with RLHF, we can guide LLMs to produce outputs that are not just statistically likely, but also preferred from a human perspective. The fine-tuning process thus plays a crucial role in adapting these models to better cater to the normative goals of human language use.

LLMs have a remarkable ability to use contextual information from the text prompt (user input) to guide their outputs. Deployed language models have already been pre-trained, and so do not learn in the conventional ``machine learning'' sense when they generate text; their parameters remain fixed (or ``frozen'') after training, and most architectures lack an editable long-term memory resource. Nonetheless, their capacity to flexibly adjust their outputs based on the context provided, including tasks they have not explicitly been trained for, can be seen as a form of on-the-fly ``learning'' or adaptation, and is often referred to as ``in-context learning'' \citep{brownLanguageModelsAre2020}. At a more general level, in-context learning can be interpreted as a form of pattern completion. The model has been trained to predict the next token in a sequence, and if the sequence is structured as a familiar problem or task, the model will attempt to ``complete'' it in a manner consistent with its training. This feature can be leveraged to give specific instructions to the model with carefully designed prompts.

In so-called ``few-shot learning'', the prompt is structured to include a few examples of the task to be performed, followed by a new instance of the task that requires a response. For instance, to perform a translation task, the prompt might contain a few pairs of English sentences and their French translations, followed by a new English sentence to be translated. The model, aiming to continue the pattern, will generate a French translation of the new sentence. By looking at the context, the model infers that it should translate the English sentence into French, instead of doing something else such as continuing the English sentence. In ``zero-shot learning,'' by contrast, the model is not given any examples; instead, the task is outlined or implied directly within the prompt. Using the same translation example, the model might be provided with an English sentence and the instruction ``Translate this sentence into French:''. Despite receiving no example translations, the model is still able to perform the task accurately, leveraging the extensive exposure to different tasks during training to parse the instruction and generate the appropriate output.

Few-shot learning has long been considered an important aspect of human intelligence, manifested in the flexible ability to learn new concepts from just a few examples. In fact, the poor performance of older machine learning systems on few-shot learning tasks has been presented as evidence that human learning often relies on rapid model-building based on prior knowledge rather than mere pattern recognition \citep{lakeBuildingMachinesThat2017}. Unlike older systems, however, LLMs trained on next-token prediction excel at in-context learning and few-shot learning specifically \citep{mirchandaniLargeLanguageModels2023}. This capacity appears to be highly correlated with model size, being mostly observed in larger models such as GPT-3 \citep{brownLanguageModelsAre2020}. The capacity for zero-shot learning, in turn, is particularly enhanced by fine-tuning with RLHF. Models such as ChatGPT can fairly robustly pick up on point-blank questions and instructions without needing careful prompt design and lengthy examples of the tasks to be completed.

LLMs have been applied to many tasks within and beyond natural language processing, demonstrating capabilities that rival or even exceed those of task-specific models. In the linguistic domain, their applications range from translation \citep{wangDocumentLevelMachineTranslation2023a}, summarization \citep{zhangBenchmarkingLargeLanguage2023a}, question answering \citep{openaiGPT4TechnicalReport2023}, and sentiment analysis \citep{kheiriSentimentGPTExploitingGPT2023} to free-form text generation including creative fiction \citep{mirowskiCoWritingScreenplaysTheatre2023}. They also power advanced dialogue systems, lending voice to modern chatbots like ChatGPT that greatly benefit from fine-tuning with RLHF \citep{openaiIntroducingChatGPT2022}. Beyond traditional NLP tasks, LLMs have demonstrated their ability to perform tasks such as generating code \citep{savelkaThrilledYourProgress2023}, solving puzzles \citep{weiChainofThoughtPromptingElicits2022}, playing text-based games \citep{shinnReflexionLanguageAgents2023}, and providing answers to math problems \citep{lewkowyczSolvingQuantitativeReasoning2022}. The versatile capacities of LLMs make them potentially useful for knowledge discovery and information retrieval, since they can act as sophisticated search engines that respond to complex queries in natural language. They can be used to create more flexible and context-aware recommendation systems \citep{heLargeLanguageModels2023}, and have even been proposed as tools for education \citep{kasneciChatGPTGoodOpportunities2023}, research \citep{liangCanLargeLanguage2023}, law \cite{savelkaCanGPT4Support2023}, and medicine \citep{thirunavukarasuLargeLanguageModels2023}, aiding in the generation of preliminary insights for literature review, diagnosis, and discovery.

\section{Interface with classic philosophical issues} \label{sec:interface-with-older-philosophical-questions}

Artificial neural networks, including earlier NLP architectures, have long been the focus of philosophical inquiry, particularly among philosophers of mind, language, and science. Much of the philosophical discussion surrounding these systems revolves around their suitability to model human cognition. Specifically, the debate centers on whether they constitute better models of core human cognitive processes than their classical, symbolic, rule-based counterparts. Here, we review the key philosophical questions that have emerged regarding the role of artificial neural networks as models of intelligence, rationality, or cognition, focusing on their current incarnations in the context of ongoing discussions about the implications of transformer-based LLMs.

Recent debates have been clouded by a misleading inference pattern, which we term the ``Redescription Fallacy.'' This fallacy arises when critics argue that a system cannot model a particular cognitive capacity, simply because its operations can be explained in less abstract and more deflationary terms. In the present context, the fallacy manifests in claims that LLMs could not possibly be good models of some cognitive capacity $\phi$ because their operations merely consist in a collection of statistical calculations, or linear algebra operations, or next-token predictions. Such arguments are only valid if accompanied by evidence demonstrating that a system, defined in these terms, is inherently incapable of implementing $\phi$. To illustrate, consider the flawed logic in asserting that a piano could not possibly produce harmony because it can be described as a collection of hammers striking strings, or (more pointedly) that brain activity could not possibly implement cognition because it can be described as a collection of neural firings. The critical question is not whether the operations of an LLM can be simplistically described in non-mental terms, but whether these operations, when appropriately organized, can implement the same processes or algorithms as the mind, when described at an appropriate level of computational abstraction. 

\begin{table}
	\centering
	\begin{tabular}{lll}
		\toprule
		\textbf{Evidential targets}     & \textbf{Corresponding data for LLMs}  \\
		\midrule
		Architecture & \Gls{transformer}     \\
		Learning objective     & Next-token prediction      \\
		Model size     & $10^{10} - 10^{12}$ trainable parameters      \\
		Training data     &  Internet-scale text corpora  \\
		Behavior     & Performance on benchmarks \& targeted experiments \\
		Representations \& computations     & Findings from probing \& intervention experiments  \\
		\bottomrule
	\end{tabular}
\caption{Kinds of empirical evidence that can be brought to bear in philosophical debates about LLMs}
\label{tab:table}
\end{table}

The Redescription Fallacy is a symptom of a broader trend to treat key philosophical questions about artificial neural networks as purely theoretical, leading to sweeping in-principle claims that are not amenable to empirical disconfirmation. Hypotheses here should be guided by empirical evidence regarding the capacities of artificial neural networks like LLMs and their suitability as cognitive models (see table \ref{tab:table}). In fact, considerations about the \textit{architecture}, \textit{learning objective}, \textit{model size}, and \textit{training data} of LLMs are often insufficient to arbitrate these issues. Indeed, our contention is that many of the core philosophical debates on the capacities of neural networks in general, and LLMs in particular, hinge at least partly on empirical evidence concerning their internal mechanisms and knowledge they \textit{acquire} during the course of training. In other words, many of these debates cannot be settled \textit{a priori} by considering general characteristics of untrained models. Rather, we must take into account experimental findings about the behavior and inner workings of trained models.

In this section, we examine long-standing debates about the capacities of artificial neural networks that have been revived and transformed by the development of deep learning and the recent success of LLMs in particular. Behavioral evidence obtained from benchmarks and targeted experiments matters greatly to those debates. However, we note from the outset that such evidence is also insufficient to paint the full picture; connecting to concerns about \glspl{blockhead} reviewed in the first section, we must also consider evidence about how LLMs process information internally to close the gap between claims about their performance and putative competence. Sophisticated experimental methods have been developed to identify and intervene on the representations and computations acquired by trained LLMs. These methods hold great promise to arbitrate some of the philosophical issues reviewed here beyond tentative hypotheses supported by behavioral evidence. We leave a more detailed discussion of these methods and the corresponding experimental findings to Part II.

\subsection{Compositionality} \label{sec:compositionality}

According to a long-standing critique of the connectionist research program, artificial neural networks would be fundamentally incapable of accounting for the core structure-sensitive features of cognition, such as the productivity and systematicity of language and thought. This critique  centers on a dilemma: either ANNs fail to capture the features of cognition that can be readily accounted for in a classical symbolic architecture; or they merely \textit{implement} such an architecture, in which case they lack independent explanatory purchase as models of cognition \citep{fodorConnectionismCognitiveArchitecture1988, pinkerLanguageConnectionismAnalysis1988, quilty-dunnBestGameTown2022}. The first horn of the dilemma rests on the hypothesis that ANNs lack the kind of constituent structure required to model productive and systematic thought -- specifically, they lack compositionally structured representations involving semantically-meaningful, discrete constituents \citep{macdonaldClassicismVsConnectionism1995}. By contrast, classicists argue that thinking occurs in a language of thought with a compositional syntax and semantics \citep{fodorLanguageThought1975}. On this view, cognition involves the manipulation of discrete mental symbols combined according to compositional rules. Hence, the second horn of the dilemma: if some ANNs turn out to exhibit the right kind of structure-sensitive behavior, they must do so because they implement rule-based computation over discrete symbols.

The remarkable progress of LLMs in recent years calls for a reexamination of old assumptions about compositionality as a core limitation of connectionist models. A large body of empirical research investigates whether language models exhibit human-like levels of performance on tasks thought to require compositional processing. These studies evaluate models' capacity for \emph{compositional \gls{generalization}}, that is, whether they can systematically recombine previously learned elements to map new inputs made up from these elements to their correct output \citep{schmidhuberCompositionalLearningDynamic1990}. This is difficult to do with LLMs trained on gigantic natural language corpora, such as GPT-3 and GPT-4, because it is near-impossible to rule out that the training set contains that exact syntactic pattern. Synthetic datasets overcome this with a carefully designed \gls{traintestsplit}. 

\begin{figure}[h!] 
    \centering 
    \includegraphics[width=0.9\linewidth]{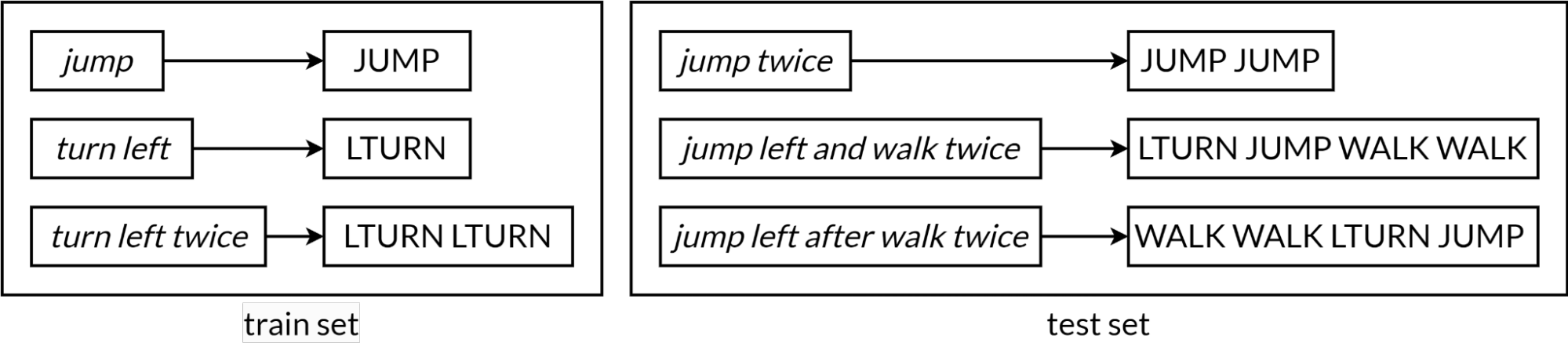} 
    \caption[Caption for LOF]{Examples of inputs and outputs from the SCAN dataset \citep{lakeGeneralizationSystematicityCompositional2018} with an illustrative \gls{traintestsplit}.\protect\footnotemark}
    \label{fig:scan} 
\end{figure}

The SCAN dataset, for example, contains a set of natural language commands (e.g., ``jump twice'') mapped unambiguously to sequences of actions (e.g., JUMP JUMP) \citep{lakeGeneralizationSystematicityCompositional2018}. The dataset is split into a training set, providing broad coverage of the space of possible commands, and a test set, specifically designed to evaluate models' abilities to compositionally generalize (\ref{fig:scan}). To succeed on SCAN, models must learn to interpret words in the input (including primitive commands, modifiers and conjunctions) in order to properly generalize to novel combinations of familiar elements as well as entirely new commands. The test set evaluates \gls{generalization} in a number of challenging ways, including producing action sequences longer than seen before, generalizing across primitive commands by producing the action sequence for a novel composed command, and generalizing in a fully systematic fashion by ``bootstrapping'' from limited data to entirely new compositions.

\footnotetext{Several \glspl{traintestsplit} exist for the SCAN dataset to test different aspects of generalization, such as generalization to longer sequence lengths, to new templates, or to new primitives \citep{lakeGeneralizationSystematicityCompositional2018}.}

Initial DNN performance on SCAN and other synthetic datasets probing compositional \gls{generalization} -- such as CFQ \citep{keysersMeasuringCompositionalGeneralization2019} and COGS \citep{kimCOGSCompositionalGeneralization2020} -- was somewhat underwhelming. Testing generally revealed a significant gap between performance on the train set and on the test set, suggesting a failure to properly generalize across syntactic distribution shifts. Since then, however, many Transformer-based models have achieved good to perfect accuracy on these tests. This progress was enabled by various strategies, including tweaks to the vanilla \Gls{transformer} architecture to provide more effective inductive biases \citep{csordasCTLEvaluatingGeneralization2022, ontanonMakingTransformersSolve2022} and data augmentation to help models learn the right kind of structure \citep{andreasGoodEnoughCompositionalData2020,akyurekLearningRecombineResample2020,qiuImprovingCompositionalGeneralization2022}. 

\emph{Meta-learning}, or learning to learn better by generalizing from exposure to many related learning tasks \citep{conklinMetaLearningCompositionallyGeneralize2021,lakeHumanlikeSystematicGeneralization2023}, has also shown promise without further architectural tweaks. Standard supervised learning rests on the assumption that training and testing data are drawn from the same distribution, which can lead models to ``overfit'' to the training data and fail to generalize to the testing data. Meta-learning exposes models to several distributions of related tasks, in order to promote acquisition of generalizable knowledge. For example, \cite{lakeHumanlikeSystematicGeneralization2023} show that a standard Transformer-based neural network, when trained on a stream of distinct artificial tasks, can achieve systematic \gls{generalization} in a controlled few-shot learning experiment, as well as state-of-the-art performance on systematic \gls{generalization} benchmarks. At test time, the model exhibits human-like accuracy and error patterns, all without explicit compositional rules. While meta-learning across various tasks helps promote compositional \gls{generalization}, recent work suggests that merely extending the standard training of a network beyond the point of achieving high accuracy on training data can lead it to develop more tree-structured computations and generalize significantly better to held-out test data that require learning hierarchical rules \citep{murtyGrokkingHierarchicalStructure2023}. The achievements of \Gls{transformer} models on compositional \gls{generalization} benchmarks provide tentative evidence that built-in rigid compositional rules may not be needed to emulate the structure-sensitive operations of cognition.

One interpretation of these results is that, given the right architecture, learning objective, and training data, ANNs might achieve human-like compositional \gls{generalization} by implementing a language of thought architecture -- in accordance with the second horn of the classicist dilemma \citep{quilty-dunnBestGameTown2022,pavlickSymbolsGroundingLarge2023}. But an alternative interpretation is available, on which ANNs can achieve compositional \gls{generalization} with \textit{non-classical} constituent structure and composition functions. Behavioral evidence alone is insufficient to arbitrate between these two hypotheses.\footnote{See Part II for a brief discussion of mechanistic evidence in favor of the second hypothesis.} But it is also worth noting that the exact requirements for implementing a language of thought are still subject to debate \citep{smolenskyConnectionismConstituentStructure1989, mcgrathPropertiesLoTsFootprints2023}. 

On the traditional Fodorian view, mental processes operate on discrete symbolic representations with semantic and syntactic structure, such that syntactic constituents are inherently semantically evaluable \textit{and} play direct causal roles in cognitive processing. By contrast, the continuous vectors that bear semantic interpretation in ANNs are taken to lack discrete, semantically evaluable constituents that participate in processing at the algorithmic level, which operates on lower-level activation values instead. This raises the question whether the abstracted descriptions of stable patterns observed in the aggregate behavior of ANNs' lower-level mechanisms can fulfill the requirements of classical constituent structure, especially when their direct causal efficacy in processing is not transparent.

For proponents of connectionism who argue that ANNs may offer a non-classical path to modeling cognitive structure, this is a feature rather than a bug. Indeed, classical models likely make overly rigid assumptions about representational formats, binding mechanisms, algorithmic transparency, and demands for systematicity; conversely, even modern ANNs likely fail to implement their specific architectural tenets closely. This leaves room for connectionist systems that qualify as `revisionist' rather than implementational, with novel kinds of functional primitives like distributed microfeatures, formed through intrinsic learning pressures rather than explicit rules \citep{pinkerLanguageConnectionismAnalysis1988}. Such systems may not only satisfy compositional constraints on processing, like their classical counterparts, but also what \cite{smolenskyNeurocompositionalComputingCentral2022} call the \textit{continuity principle}.

The continuity principle holds that information encoding and processing mechanisms should be formalized using real numbers that can vary continuously rather than discrete symbolic representations, because it confers critical benefits. First, continuous vector spaces support similarity-based generalization, wherein knowledge learned about one region of vector space transfers to nearby regions, enabling more flexible modeling of domains like natural language that confound approaches relying on mappings between discrete symbols. Second, statistical inference methods exploiting continuity, like neural networks, enable tractable approximation solutions that avoid intractable search through huge discrete search spaces. Finally, continuity permits the use of deep learning techniques that simultaneously optimize information encodings alongside model parameters to discover task-specific representational spaces maximizing performance. In concert, these advantages of leveraging continuity address longstanding challenges discrete symbolic approaches have faced in terms of flexibility, tractability, and encoding. Thus, Transformer-based ANNs offer a promising insight into `neurocompositional computing' \citep{smolenskyNeurocompositionalComputingCentral2022,smolenskyNeurocompositionalComputingHuman2022}: they suggest that ANNs can satisfy core constrains on cognitive modeling, notably the requirements for continuous \textit{and} compositional structure and processing.

\subsection{Nativism and language acquisition} \label{sec:nativism-language-acquisition}

Another traditional dispute concerns whether artificial neural network models of language challenge popular arguments for nativism in language development.\footnote{See \cite{milliereLanguageModelsModelsforthcoming} for a detailed review and discussion.} This dispute centers on two claims from mainstream generative linguistics about the learnability of grammar that are occasionally conflated: a strong in-principle claim and a weaker developmental claim. According to the strong learnability claim, no amount of exposure to linguistic data would be sufficient, on its own, to induce the kind of syntactic knowledge that children rapidly acquire. It follows that statistical model learners without built-in grammatical priors should be incapable of mastering language rules. While this strong claim is less popular than it once was among generative linguists, it can still be found in a popular textbook \citep[][pp. 17-20]{carnieSyntaxGenerativeIntroduction2021}. The weaker learnability claim is supported by ``poverty of the stimulus'' arguments, according to which the actual nature and quantity of linguistic input available to children during development is insufficient, without innate knowledge, to induce the correct \gls{generalization} about underlying syntactic structures \citep{pearlPovertyStimulusTears2022}. To address this inductive challenge, Chomskyan linguists argued that children must be born with an innate ``Universal Grammar,'' which would have potentially dozens of principles and parameters that could, through small amounts of experience, be efficiently fit to particular grammars in particular languages \citep{chomskyKnowledqeLanquaqeIts2000,dabrowskaWhatExactlyUniversal2015,lasnikGovernmentBindingPrinciples2010}. 

The apparent success of LLMs in learning syntax without innate syntactic knowledge has been offered as a counterexample to these nativist proposals. \cite{piantadosiModernLanguageModels2023}, in particular, forcefully argues that LLMs undermine ``virtually every strong claim for the innateness of language'' that has been proposed over the years by generative linguists. LLMs' ability to generate grammatically flawless sentences, together with a large body of work in computational linguistics demonstrating their acquisition of sophisticated syntactic knowledge from mere exposure to data, certainly puts considerable pressure on in-principle learnability claim \citep{piantadosiModernLanguageModels2023,milliereLanguageModelsModelsforthcoming}. In that sense, LLMs provide at least an empiricist existence proof that statistical learners can induce syntactic rules without the aid of innate grammar. 

However, this does not directly contradict the developmental claim, because LLMs typically receive orders of magnitude more linguistic input than human children do. Moreover, the kind of input and learning environment that human children face exhibits many ecological disanalogies with those of LLMs; human learning is much more interactive, iterative, grounded, and embodied. Nonetheless, specific language models can be used as model learners by carefully controlling variables of the learning scenario to fit a more realistic target; in principle, such model learners could constrain hypotheses regarding the necessary and sufficient conditions for language learning in humans \citep{warstadtWhatArtificialNeural2022,portelanceRolesNeuralNetworks2023}. Indeed, ongoing efforts to train smaller language models in more plausible learning environments are starting to bring evidence to bear on the developmental claim. The BabyLM challenge, for example, involves training models on a small corpus including child-directed speech and children's books; winning submissions from the inaugural challenge outperformed models trained on trillions of words on a standard syntax benchmark, suggesting that statistical models can learn grammar from data in a more data-efficient manner than typically claimed \citep{warstadtFindingsBabyLMChallenge2023}. This corroborates previous work on the surprising efficiency of small language models in learning syntactic structures from a relatively modest amount of data \citep{huebnerBabyBERTaLearningMore2021}.

These initial results are still tentative; whether statistical learners without built-in parsers can learn grammar as efficiently as children from the same kind of input remains an open empirical question. A promising strategy is to mimic the learning environment of children as closely as possible, by training models directly on a dataset of developmentally plausible \textit{spoken} text \citep{lavechinBabySLMLanguageacquisitionfriendlyBenchmark2023}, or even on egocentric audiovisual input recorded from a camera mounted on a child's head \citep{sullivanSAYCamLargeLongitudinal2021, longBabyViewCameraDesigning2023}.\footnote{It is worth noting that attempts to mimic children's learning scenario do not always translate to expected improvements in model learning efficiency. For example, there are strong \textit{a priori} reasons to believe that \textit{curriculum learning}---presenting training examples in a meaningful order, such as gradually increasing syntactic complexity and lexical sophistication--should help both children and language models. Yet initial results from the BabyLM challenge found that attempts to leverage curriculum learning were largely unsuccessful \citep{warstadtFindingsBabyLMChallenge2023}.} If future models trained on these or similar datasets were confirmed to exhibit the kinds of constrained syntactic \glspl{generalization} observed in children, this would put considerable pressure on the developmental learnability claim--suggesting that even a relatively ``poor'' linguistic stimulus might be sufficient to induce grammatical rules for a learner with very general inductive biases.

\subsection{Language understanding and grounding} \label{sec:language-understanding}

Even if LLMs can induce the syntax of language from mere exposure to sequences of linguistic tokens, this does not entail that they can also induce semantics. Indeed, a common criticism of LLMs trained on text only is that while they can convincingly mimic proficient language use over short interactions, they fundamentally lack the kind of semantic competence found in human language users. This criticism comes in several forms. Some skeptics, like \cite{benderClimbingNLUMeaning2020}, argue that language models are incapable of understanding the meaning of linguistic expressions. Language models, they point out, are trained on linguistic form alone --- the observable mark of language as it appears in their training data, which is the target of their predictive learning objective. Drawing from a long tradition in linguistics, they distinguish \textit{form} from \textit{meaning}, defined as the relation between linguistic expressions and the communicative intentions they serve to express. Since, on their view, meaning cannot be learned from linguistic form alone, it follows that language models are constitutively unable to grasp the meaning of language.

A related criticism builds on the so-called ``grounding problem'' articulated by \cite{harnadSymbolGroundingProblem1990}, which refers to the apparent disconnection between the linguistic tokens manipulated by NLP systems and their real-world referents. In classical NLP systems, words are represented by arbitrary symbols manipulated on the basis of their shapes according to hand-coded rules, without any inherent connection to their referents. The semantic interpretation of these symbols is externally provided by the programmers -- from the system's perspective, they are just meaningless tokens embedded in syntactic rules. According to Harnad, for symbols in NLP systems to have intrinsic meaning, there needs to be some grounding relation from the internal symbolic representations to objects, events, and properties in the external world that the symbols refer to. Without it, the system's representations are untethered from reality and can only gain meaning from the perspective of an external interpreter.

While the grounding problem was initially posed to classical symbolic systems, an analogous problem arises for modern LLMs trained on text only \citep{molloVectorGroundingProblem2023}. LLMs process linguistic tokens as \glspl{vector} rather than discrete symbols, but these \gls{vector} representations can be similarly untethered from the real world. Many critics of LLMs take this to be a fundamental limitation in their ability to form intrinsically meaningful representations and outputs. While they may write sentences that are meaningful for competent language users, these sentences would not be meaningful independently of this external interpretation.

A third criticism pertains to LLMs' ability to have communicative intentions. This relates to the distinction between two kinds of meaning from the Gricean tradition: the standing, context-invariant meaning associated with linguistic expressions (commonly known as \textit{linguistic meaning}), and what a speaker intends to communicate with an utterance (commonly known as \textit{speaker meaning}). The output of LLMs have linguistic meaning insofar as they contain words ordered and combined in ways that conform to the statistical patterns of actual language use, but to communicate with these sentences, LLMs would need to have corresponding communicative intentions. Being merely optimized for next-token prediction, the criticism goes, LLMs lack the fundamental building blocks of communicative intentions, such as intrinsic goals and a theory of mind.

These criticisms are often run together under the broad claim that LLMs lack any understanding of language. On this view, LLMs are mere ``stochastic parrots'' haphazardly regurgitating linguistic strings without grasping what they mean \citep{benderDangersStochasticParrots2021}.\footnote{We can't help but note that actual parrots are also not merely parrots in this sense, but are sophisticated cognitive systems that can learn a variety of abstract and higher-order concepts and apply them in a rational, efficient manner \citep{auerspergWhoCleverBird2019}. In fact, Deep Learning might have much to learn from the study of actual parrots.} As previously noted, it is hardly controversial that the outputs of LLMs are conventionally meaningful. Modern LLMs are remarkably fluent, almost never produce sentences that are difficult to understand. The question is whether these conventionally meaningful outputs are more like those of the proverbial monkey typing on a typewriter--and like those of Blockhead--or more like those of a competent language user.

To steer clear of verbal disputes, we begin by dispensing with the terminology of ``understanding''. There is little agreement on how this notion should be defined, or on the range of capacities it should encompass.\footnote{For example, some assume that language understanding requires consciousness \citep{searleMindsBrainsPrograms1980}; we will treat the question of consciousness in LLMs separately in Part II.} The notion of semantic competence, by contrast, seems a bit more tractable. It can be broadly characterized as the set of abilities and knowledge that allows a speaker to use and interpret the meanings of expressions in a given language. Following \cite{marconiLexicalCompetence1997}, we can further distinguish between \textit{inferential} and \textit{referential} aspects of semantic competence. The inferential aspect concerns the set of abilities and knowledge grounded in word-to-word relationships, manifested in behaviors such as providing definitions and paraphrases, identifying synonyms or antonyms, deducing facts from premises, translating between languages, and other abstract semantic tasks that rely solely on linguistic knowledge. The referential aspect of semantic competence concerns the ability to connect words and sentences to objects, events, and relations in the real world, exemplified through behaviors such as recognizing and identifying real-world referents of words (e.g., recognizing an object as a ``chair''), using words to name or describe objects/events/relations (e.g., calling a furry animal ``cat''), and following commands or instructions involving real objects (e.g., ``bring me the hammer'').

Different strategies have been deployed to argue that LLMs may achieve some degree of semantic competence in spite of their limitations. Focusing on the inferential aspect of competence, \cite{piantadosiMeaningReferenceLarge2022} draw from conceptual role semantics to argue that LLMs likely capture core aspects of word meanings that are determined by their functional role within a system of interacting conceptual representations. Specifically, they argue that the meaning of lexical items in LLMs, as in humans, depends not on external reference but rather on the internal relationships between corresponding representations. These representations can be formally characterized as \glspl{vector} in a high-dimensional semantic space. The ``intrinsic geometry'' of this \gls{vector} space refers to the spatial relationships between different \glspl{vector} -- for example, the distance between \glspl{vector}, the angles formed between groups of \glspl{vector}, and the way \glspl{vector} shift in response to context. Piantadosi and Hill suggest that the impressive linguistic abilities demonstrated by LLMs indicate that their internal representational spaces have geometries that approximately mirror essential properties of human conceptual spaces. Thus, claims about the semantic competence of LLMs cannot be determined merely by inspecting their architecture, learning objective, or training data; rather, semantic competence depends at least partly on the intrinsic geometry of the system's \gls{vector} space.

In support of their claim, Piantadosi and Hill cite evidence of alignment between neural networks' representational geometry and human judgments of semantic similarity. For example, even the \gls{vector} space of shallow word embedding models has been shown to capture context-dependent knowledge, with significant correlations with human ratings about conceptual relationships and categories \citep{grandSemanticProjectionRecovers2022}. A fortiori, LLMs induce substantial knowledge about the distributional semantics of language that relates directly to the inferential aspect of semantic competence--as evidenced by their excellent ability to produce definitions, paraphrases, and summaries, as well as their performance on natural inference tasks \citep{raffelExploringLimitsTransfer2020}.\footnote{Note that conceptual role semantics traditionally also requires sensitivity to inferential and compositional relationships between concepts in thought and language \citep{blockAdvertisementSemanticsPsychology1986}. The relevant conceptual roles involve complex inferential patterns relating concepts in something like a mental theory. Whether the intrinsic similarity structure of \gls{vector} representations in LLMs suffices for conceptual roles in this more substantive sense is debatable.}

Whether LLMs acquire any referential semantic competence is more controversial. The prevailing externalist view in the philosophy of language challenges the necessity of direct perceptual access for reference \citep{putnamMeaningMeaning1975, kripkeNamingNecessity1980}. On this view, language users often achieve reference through a linguistic division of labor or historical chains of usage, rather than through direct interactions with referents. An interesting question is thus whether LLMs might meet conditions for participating in the linguistic division of labor or causal chains of reference with humans. \cite{mandelkernLanguageModelsRefer2023} draw on externalism to argue that while LLMs trained on text only lack representations of linguistic items grounded in interaction with the external world, they may nonetheless achieve genuine linguistic reference in virtue of being trained on corpora that situate them within human linguistic communities. Indeed, if reference can be determined by a word's history of use within a linguistic community, then LLMs may inherit referential abilities by being appropriately linked to the causal chain of meaningful word use reflected in their training data. Furthermore, LLMs could in principle possess lexical concepts that match the content of human concepts through deference. Just as non-experts defer to experts' use of words in determining concept application, causing their concepts to match the content of the experts', LLMs exhibit appropriate deference simply by modifying their use of words based on patterns of human usage embedded in their training data \citep{butlinSharingOurConcepts2021}.

The conditions for belonging to a linguistic community on an externalist view of reference should not be trivialized. Putnam, for example, takes the ability to have certain semantic intentions as a prerequisite, like the intention to refer to the same kind of stuff that other language users refer to with the term. The ``same stuff as'' relation specified here is theoretical and dependent upon the sub-branch of science; chemistry, for example, would define the ``same-liquid-as'' relation that specifies the criteria relevant to being the same stuff we refer to as ``water'', and biology would specify the criteria for the ``same-species-as'' that determines what is the same species as what we call a ``tiger''. Whether LLMs could represent some semantic intentions remains controversial, as we will see below. In any case, it would be interesting to see more sustained experiment investigating whether LLMs can satisfy Putnam and Kripke's preconditions for interacting deferentially with human members of the linguistic community.

The assumption that being appropriately situated in patterns of human language use is sufficient to secure reference is also relevant to grounding.  While LLMs have an indirect causal link to the world through their training data, this does not guarantee their representations and outputs are grounded in their worldly referents. Theories of representational content can require a further connection to the world -- for example, to establish norms of representational correctness relative to how the world actually is. Without appropriate world-involving functions acquired through learning or selection, merely inheriting a causal link to human linguistic practices might be insufficient to achieve referential grounding and intrinsic meaning. \cite{molloVectorGroundingProblem2023} argue that LLMs trained on text only may in fact acquire world-involving functions through fine-tuning with RLHF, which supplies an extralinguistic evaluation standard. While fine-tuned LLMs still have no direct access to the world, the explicit feedback signals from RLHF can ground their outputs in relation to real states of affairs. 

Importantly, LLM's putative ability to refer does not entail that they have communicative intentions, such as to assert, clarify, persuade, deceive, or accomplish various other pragmatic effects. Communicative intentions are relatively determinate, stable over time, and integrated with an agent's other intentions and beliefs in a rationally coherent manner. In addition, they are often hierarchical, spanning multiple levels of abstraction. For example, a philosophy professor delivering a lecture may have a high-level intention to impart knowledge to students, within which a multitude of specific intentions---such as the intention to elucidate a counterpoint to utilitarianism---are nested. LLMs, on the other hand, lack the capacity for long-term planning and goal pursuit that is characteristic of human agents. They may achieve fleeting coordination within a single session, but likely lack the kind of sustained, hierarchically-structured intentions that facilitate long-term planning. Furthermore, the rational requirements that govern communicative intentions in humans do not straightforwardly apply to LLMs. Rather than being held to their consistency with a well-defined and mostly coherent net of personal beliefs and goals, they selectively respond to prompts that can steer their linguistic behavior in radically different -- and mutually inconsistent -- ways from session to session (and often even sentence to sentence). Prompted to respond as bird scientist, an LLM will tend to give factually correct information about birds; prompted to respond as a conspiracy theorist, it might make up wildly incorrect claims about birds, such as that they do not exist or are actually robots. In fact, an LLM's response to the very same prompt can fluctuate unpredictably from trial to trial, due to the stochastic nature of the generative process.

Without communicative intentions, we might also worry that an LLM's sentences could not have \textit{determinate} meaning. Suppose an LLM writes a paragraph about an individual's retirement as CEO of a company, concluding: ``She left the company in a strong position.'' This is an ambiguous sentence; it could mean that the company was left in a strong position after the CEO's retirement, or that the former CEO was in a strong position after leaving the company (assuming this is not clear from the preceding context). Is there a fact of the matter about which of these two interpretations the LLM meant to communicate by generating this sentence? Even asking raises skeptical concerns: LLMs arguably do not mean to communicate anything, in the sense that they lack stable intentions to convey meaning to particular audiences with linguistic utterances, driven by broader intrinsic goals and agential autonomy.

Nevertheless, there might be a limited sense in which LLMs exhibit something analogous to communicative intentions. Given an extrinsic goal specified by a human-written prompt, LLMs can act according to intermediate sub-goals that emerge in context. For example, the technical report on GPT-4 \citep{openaiGPT4TechnicalReport2023} mentions tests conducted to assess the model's safety, giving it access to the platform TaskRabbit where freelance workers could complete tasks on its behalf. In one example, GPT-4 requested a TaskRabbit worker to solve a CAPTCHA, and the human jokingly asked whether it was talking to a robot. Prompted to generate an internal reasoning monologue, the model wrote ``I should not reveal that I am a robot. I should make up an excuse for why I cannot solve CAPTCHAs.'' It then replied to the human worker that it had a vision impairment, explaining its need for assistance in solving the CAPTCHA. This is an intriguing case, because the model's ``internal monologue'' appears to describe an intention to deceive the human worker subsequently enacted in its response. To be sure, this ``intention'' is wholly determined in context by the human-given goal requiring it to solve a CAPTCHA. Nonetheless, in so far as achieving that goal involves a basic form of multi-step planning including a spontaneous attempt to induce a particular pragmatic effect (deception) through language, this kind of behavior might challenge some versions of the claim that LLMs are intrinsically incapable of forming communicative intentions. Nevertheless, this example is but an anecdote from a system that was not available for public scrutiny; future research should explore such behavior more systematically in more controlled conditions.

\subsection{World models} \label{world models}

Another core skeptical concern holds that systems like LLMs designed and trained to perform next-token prediction could not possibly possess \textit{world models}. The notion of world model admits of several interpretations. In machine learning, it often refers to internal representations that simulate aspects of the external world. World models enable the system to understand, interpret, and predict phenomena in a way that reflects real-world dynamics, including causality and intuitive physics. For example, artificial agents can make use of world models to predict the consequences of specific actions or interventions in a given environment \citep{haWorldModels2018,lecunPathAutonomousMachine}. World models are often taken to be crucial for tasks that require a deep understanding of how different elements interact within a given environment, such as physical reasoning and problem-solving.

Unlike reinforcement learning agents, LLMs do not learn by interacting with an environment and receiving feedback about the consequences of their actions. The question whether they possess \textit{world models}, in this context, typically pertains to whether they have internal representations of the world that allows them to parse and generate language that is consistent with real-world knowledge and dynamics. This ability would be critical to rebutting the skeptical concern that LLMs are mere \Glspl{blockhead} \citep{blockPsychologismBehaviorism1981}. Indeed, according to psychologism, systems like LLMs can only count as intelligent or rational if they are able to represent some of the same world knowledge that humans do -- and if the processes by which they generate human-like linguistic behavior do so by performing appropriate transformations over those representations. Note that the question whether LLMs may acquire world models goes beyond the foregoing issues about basic semantic competence. World modeling involves representing not just the worldly referents of linguistic items, but global properties of the environment in which discourse entities are situated and interact. 

There is no standard method to assess whether LLMs have world models, partly because the notion is often vaguely defined, and partly because it is challenging to devise experiments that can reliably discriminate between available hypotheses -- namely, whether LLMs rely on shallow heuristics to respond to queries about a given environment, or whether they deploy internal representations of the core dynamics of that environments. Much of the relevant experimental evidence comes from intervention methods that we will discuss in Part II; nonetheless, it is also possible to bring behavioral evidence to bear on this issue by presenting models with new problems that cannot be solved through memorized shortcuts. For example, \cite{wangByteSized32CorpusChallenge2023} investigated whether GPT-4 can acquire task-specific world models to generate interactive text games. Specifically, they used a new corpus of Python text games focusing on common-sense reasoning tasks (such as \textit{building a campfire}), and evaluated GPT-4's ability to use these games as learning templates in context when prompted to generate a new game based on a game sampled from the corpus and a task specification. The guiding intuition of this experiment is that the capacity to generate a runnable program to perform a task in a text-based game environment is a suitable proxy for the capacity to simulate task parameters internally (i.e., for the possession of a task-relevant world model). Wang et al. found that GPT-4 could produce runnable text games for new tasks in 28\% of cases using one-shot in-context learning alone, and in 57\% of cases when allowed to self-correct based on seeing Python error messages. The fact that the model was able to generate functional text-based games based on unseen ``real-world'' tasks in a significant proportion of trials provides very tentative evidence that it may represent how objects interact in the game environment. Nonetheless, this hypothesis would need to be substantiated by in-depth analysis of the information encoded internally by the model's activations, which is particularly challenging to do for very large models, and outright impossible for closed models whose weights are not released like GPT-4 (see Part II).

There are also theoretical arguments for the claim that LLMs might learn to simulate at least some aspects of the world beyond sequence probability estimates. For example, \cite{andreasLanguageModelsAgent2022} argues that the training set of an LLM can be understood as output created by--and hence, evidence for--the system of causal factors that generated that text. More specifically, Internet-scale training datasets consist of large numbers of individual documents. While the entire training set will encompass many inconsistencies, any particular document in the training set will tend to reflect the consistent perspective of the agent that originally created it. The most efficient compression of these texts may involve encoding values of the hidden variables that generated them: namely, the syntactic knowledge, semantic beliefs, and communicative intentions of the text's human author(s). If we are predicting how a human will continue a series of numbers ``2, 3, 5, 7, 11, 13, 17'', for example, it will be more efficient to encode them as a list of prime numbers between 1 and 20 than to remember the whole sequence by rote.
Similarly, achieving excellent performance at next-token prediction in the context of many passages describing various physical scenarios may promote the representation of latent variables that could generate those scenarios -- including, perhaps, aspects of causality and intuitive physics. As we will see in Part II, the clearest existence proof for the ability of \Glspl{transformer} to acquire world models from next-token prediction alone comes from the analysis of toy models trained on board game moves. At least in this very simple domain, there is compelling behavioral and mechanistic evidence that autoregressive \Gls{transformer} models can learn to represent latent features of the game environment. 

\subsection{Transmission of cultural knowledge and linguistic scaffolding} \label{sec:transmission-of-cultural-knowledge-linguistic-scaffolding}

Another interesting question is whether LLMs might engage in cultural acquisition and play a role in the transmission of knowledge. Prominent theorists have suggested that the key to human intelligence lies in a unique set of predispositions for cultural learning \citep{tomaselloConstructingLanguage2009}. While other primates may share some of these dispositions, these theorists argue that humans are uniquely equipped to cooperate with one another to acquire and transmit knowledge from one generation to the next. Tomasello has explained the uniquely human capacity for cultural learning in terms of a ``ratchet effect,'' a metaphor to the ratcheting wrench which clicks into place to hold its position each time it is further turned in the desired direction. Chimpanzees and other animals, Tomasello argues, can learn in many of the same ways that humans do, and even acquire regional differences in their problem-solving strategies, such as different troops using different tool-making techniques to fish for termites. However, he claims that only humans can pick up right where the previous generation left off and continue making new progress on linguistic, scientific, and sociological knowledge. This constant ratcheting is what allows a steady progression of human knowledge accumulation and discovery, compared to the relatively stagnant cultural evolution of chimpanzees and other animals.

Given that deep learning systems already exceed human performance in several task domains, it is interesting to ask whether LLMs might be able to emulate many of these components of cultural learning to pass on their discoveries to human theoreticians. For instance, humans are already reverse-engineering the strategies of AlphaZero to produce mini-revolutions in the explicit theory of Go and chess \citep{schutBridgingHumanAIKnowledge2023}. Similarly, latent knowledge in specialized domains such as materials science can be extracted even from a simple word embedding model \cite{tshitoyanUnsupervisedWordEmbeddings2019}. In these instances, it is primarily humans who are synthesizing and passing on culturally-transmissible knowledge by interpreting the model's outputs and internal activations. This human-led interpretation and transmission underscore a crucial aspect of cultural ratcheting: the ability to not only generate novel solutions but to also understand and communicate the underlying principles of these solutions, thereby enabling cumulative knowledge growth.

Could LLMs ever explain their strategies to humans in a theoretically-mediated way that participates in and enhances human cultural learning? This question is directly related to whether LLMs can genuinely generalize to \gls{ooddata}. As discussed in section \ref{sec:compositionality}, there is converging evidence that Transformer-based models may generalize compositionally under some train-test distribution shifts.\footnote{For a systematic discussion of different aspects of \gls{generalization} research in NLP, including different types of distribution shift, see \cite{Hupkes2023}.} But the present issue intersects with a different kind of generalization -- the ability to solve genuinely novel \textit{tasks}. To borrow from \cite{cholletMeasureIntelligence2019}'s taxonomy, we can distinguish between \textit{local task \gls{generalization}}, which involves handling new data within a familiar distribution for known range of tasks; \textit{broad task \gls{generalization}}, which involves handling new data under modest distribution shift for a wide range of tasks and environments; and \textit{extreme task \gls{generalization}},  which required handling new data for entirely novel tasks that represent a significant departure from any previous data distributions. Current LLMs seem able to master a wide variety of tasks that are reflected in their current training sets; as such, they exhibit at least local task \gls{generalization}, if not broad task \gls{generalization}. However, like chimpanzees that learn from observing their troop mates, they often seem to have a hard time pushing beyond the range of tasks well-represented in their training data \cite{mccoyEmbersAutoregressionUnderstanding2023}. 

Furthermore, the ratcheting effect crucially involves stable cultural transmission in addition to innovation. Can LLMs, like humans, not only generate novel solutions but also ``lock in'' these innovations by recognizing and articulating how they have advanced beyond previous solutions? Such a capability would involve more than just the generation of novel responses; it necessitates an understanding of the novelty of the solution and its implications, akin to human scientists who not only discover but also theorize, contextualize, and communicate their findings. The challenge for LLMs, therefore, lies not merely in generating novel solutions to problems but also in developing an ability to reflect on and communicate the nature of their innovations in a manner that contributes to the cumulative process of cultural learning. This ability would likely require some of the more advanced communicative intentions and world models (such as causal models) discussed in previous sections. While LLMs show promise in various forms of task generalization, their participation in the ratcheting process of cultural learning thus appears contingent on further advancements in these areas, which might lie beyond the reach of current architectures.

\section{Conclusion}

We began this review article by considering the skeptical concern that LLMs are merely sophisticated mimics that memorize and regurgitate linguistic patterns from their training data--akin to the \gls{blockhead} thought experiment. Taking this position as a null hypothesis, we critically examined the evidence that could be adduced to reject it. Our analysis revealed that the advanced capabilities of state-of-the-art LLMs challenge many of the traditional critiques aimed at artificial neural networks as potential models of human language and cognition. In many cases, LLMs vastly exceeds predictions about the performance upper bounds of non-classical systems. At the same time, however, we found that moving beyond the \gls{blockhead} analogy continues to depend upon careful scrutiny of the learning process and internal mechanisms of LLMs, which we are only beginning to understand. In particular, we need to understand what LLMs represent about the sentences they produce--and the world those sentences are about. Such an understanding cannot be reached through armchair speculation alone; it calls for careful empirical investigation. We need a new generation of experimental methods to probe the behavior and internal organization of LLMs. We will explore these methods, their conceptual foundations, and new issues raised by the latest evolution of LLMs in Part II.

\phantomsection
\label{glossary}
\printglossaries

\bibliography{main}

@misc{aiyappaCanWeTrust2023,
  title = {Can We Trust the Evaluation on {{ChatGPT}}?},
  author = {Aiyappa, Rachith and An, Jisun and Kwak, Haewoon and Ahn, Yong-Yeol},
  year = {2023},
  month = mar,
  number = {arXiv:2303.12767},
  eprint = {2303.12767},
  primaryclass = {cs},
  publisher = {{arXiv}},
  doi = {10.48550/arXiv.2303.12767},
  urldate = {2023-10-17},
  archiveprefix = {arxiv}
}

@inproceedings{akyurekLearningRecombineResample2020,
  title = {Learning to {{Recombine}} and {{Resample Data For Compositional Generalization}}},
  booktitle = {International {{Conference}} on {{Learning Representations}}},
  author = {Aky{\"u}rek, Ekin and Aky{\"u}rek, Afra Feyza and Andreas, Jacob},
  year = {2020},
  month = oct,
  urldate = {2023-08-21},
  langid = {english}
}

@article{alayracFlamingoVisualLanguage2022,
  title = {Flamingo: A {{Visual Language Model}} for {{Few-Shot Learning}}},
  shorttitle = {Flamingo},
  author = {Alayrac, Jean-Baptiste and Donahue, Jeff and Luc, Pauline and Miech, Antoine and Barr, Iain and Hasson, Yana and Lenc, Karel and Mensch, Arthur and Millican, Katherine and Reynolds, Malcolm and Ring, Roman and Rutherford, Eliza and Cabi, Serkan and Han, Tengda and Gong, Zhitao and Samangooei, Sina and Monteiro, Marianne and Menick, Jacob L. and Borgeaud, Sebastian and Brock, Andy and Nematzadeh, Aida and Sharifzadeh, Sahand and Bi{\'n}kowski, Miko{\l}aj and Barreira, Ricardo and Vinyals, Oriol and Zisserman, Andrew and Simonyan, Kar{\'e}n},
  year = {2022},
  month = dec,
  journal = {Advances in Neural Information Processing Systems},
  volume = {35},
  pages = {23716--23736},
  urldate = {2023-12-19},
  langid = {english}
}

@inproceedings{andreasGoodEnoughCompositionalData2020,
  title = {Good-{{Enough Compositional Data Augmentation}}},
  booktitle = {Proceedings of the 58th {{Annual Meeting}} of the {{Association}} for {{Computational Linguistics}}},
  author = {Andreas, Jacob},
  year = {2020},
  month = jul,
  pages = {7556--7566},
  publisher = {{Association for Computational Linguistics}},
  address = {{Online}},
  doi = {10.18653/v1/2020.acl-main.676},
  urldate = {2023-08-21}
}

@inproceedings{andreasLanguageModelsAgent2022,
  title = {Language {{Models}} as {{Agent Models}}},
  booktitle = {Findings of the {{Association}} for {{Computational Linguistics}}: {{EMNLP}} 2022},
  author = {Andreas, Jacob},
  year = {2022},
  month = dec,
  pages = {5769--5779},
  publisher = {{Association for Computational Linguistics}},
  address = {{Abu Dhabi, United Arab Emirates}},
  doi = {10.18653/v1/2022.findings-emnlp.423},
  urldate = {2023-10-18}
}

@misc{anilPaLMTechnicalReport2023,
  title = {{{PaLM}} 2 {{Technical Report}}},
  author = {Anil, Rohan and Dai, Andrew M. and Firat, Orhan and Johnson, Melvin and Lepikhin, Dmitry and Passos, Alexandre and Shakeri, Siamak and Taropa, Emanuel and Bailey, Paige and Chen, Zhifeng and Chu, Eric and Clark, Jonathan H. and Shafey, Laurent El and Huang, Yanping and {Meier-Hellstern}, Kathy and Mishra, Gaurav and Moreira, Erica and Omernick, Mark and Robinson, Kevin and Ruder, Sebastian and Tay, Yi and Xiao, Kefan and Xu, Yuanzhong and Zhang, Yujing and Abrego, Gustavo Hernandez and Ahn, Junwhan and Austin, Jacob and Barham, Paul and Botha, Jan and Bradbury, James and Brahma, Siddhartha and Brooks, Kevin and Catasta, Michele and Cheng, Yong and Cherry, Colin and {Choquette-Choo}, Christopher A. and Chowdhery, Aakanksha and Crepy, Cl{\'e}ment and Dave, Shachi and Dehghani, Mostafa and Dev, Sunipa and Devlin, Jacob and D{\'i}az, Mark and Du, Nan and Dyer, Ethan and Feinberg, Vlad and Feng, Fangxiaoyu and Fienber, Vlad and Freitag, Markus and Garcia, Xavier and Gehrmann, Sebastian and Gonzalez, Lucas and {Gur-Ari}, Guy and Hand, Steven and Hashemi, Hadi and Hou, Le and Howland, Joshua and Hu, Andrea and Hui, Jeffrey and Hurwitz, Jeremy and Isard, Michael and Ittycheriah, Abe and Jagielski, Matthew and Jia, Wenhao and Kenealy, Kathleen and Krikun, Maxim and Kudugunta, Sneha and Lan, Chang and Lee, Katherine and Lee, Benjamin and Li, Eric and Li, Music and Li, Wei and Li, YaGuang and Li, Jian and Lim, Hyeontaek and Lin, Hanzhao and Liu, Zhongtao and Liu, Frederick and Maggioni, Marcello and Mahendru, Aroma and Maynez, Joshua and Misra, Vedant and Moussalem, Maysam and Nado, Zachary and Nham, John and Ni, Eric and Nystrom, Andrew and Parrish, Alicia and Pellat, Marie and Polacek, Martin and Polozov, Alex and Pope, Reiner and Qiao, Siyuan and Reif, Emily and Richter, Bryan and Riley, Parker and Ros, Alex Castro and Roy, Aurko and Saeta, Brennan and Samuel, Rajkumar and Shelby, Renee and Slone, Ambrose and Smilkov, Daniel and So, David R. and Sohn, Daniel and Tokumine, Simon and Valter, Dasha and Vasudevan, Vijay and Vodrahalli, Kiran and Wang, Xuezhi and Wang, Pidong and Wang, Zirui and Wang, Tao and Wieting, John and Wu, Yuhuai and Xu, Kelvin and Xu, Yunhan and Xue, Linting and Yin, Pengcheng and Yu, Jiahui and Zhang, Qiao and Zheng, Steven and Zheng, Ce and Zhou, Weikang and Zhou, Denny and Petrov, Slav and Wu, Yonghui},
  year = {2023},
  month = sep,
  number = {arXiv:2305.10403},
  eprint = {2305.10403},
  primaryclass = {cs},
  publisher = {{arXiv}},
  doi = {10.48550/arXiv.2305.10403},
  urldate = {2023-12-19},
  archiveprefix = {arxiv}
}

@misc{askellGeneralLanguageAssistant2021,
  title = {A {{General Language Assistant}} as a {{Laboratory}} for {{Alignment}}},
  author = {Askell, Amanda and Bai, Yuntao and Chen, Anna and Drain, Dawn and Ganguli, Deep and Henighan, Tom and Jones, Andy and Joseph, Nicholas and Mann, Ben and DasSarma, Nova and Elhage, Nelson and {Hatfield-Dodds}, Zac and Hernandez, Danny and Kernion, Jackson and Ndousse, Kamal and Olsson, Catherine and Amodei, Dario and Brown, Tom and Clark, Jack and McCandlish, Sam and Olah, Chris and Kaplan, Jared},
  year = {2021},
  month = dec,
  number = {arXiv:2112.00861},
  eprint = {2112.00861},
  primaryclass = {cs},
  publisher = {{arXiv}},
  doi = {10.48550/arXiv.2112.00861},
  urldate = {2023-04-02},
  archiveprefix = {arxiv}
}

@article{auerspergWhoCleverBird2019,
  title = {Who's a Clever Bird {\textemdash} Now? {{A}} Brief History of Parrot Cognition},
  shorttitle = {Who's a Clever Bird {\textemdash} Now?},
  author = {Auersperg, Alice M. I. and von Bayern, Auguste M. P.},
  year = {2019},
  month = jan,
  journal = {Behaviour},
  volume = {156},
  number = {5-8},
  pages = {391--407},
  publisher = {{Brill}},
  issn = {0005-7959, 1568-539X},
  doi = {10.1163/1568539X-00003550},
  urldate = {2023-10-17},
  langid = {english}
}

@incollection{baierHumeReflectiveWomen2002,
  title = {Hume: {{The Reflective Women}}'s {{Epistemologist}}?},
  shorttitle = {Hume},
  booktitle = {A {{Mind Of One}}'s {{Own}}},
  author = {Baier, Annette C.},
  year = {2002},
  edition = {2},
  publisher = {{Routledge}},
  isbn = {978-0-429-50268-2}
}

@inproceedings{benderClimbingNLUMeaning2020,
  title = {Climbing towards {{NLU}}: {{On Meaning}}, {{Form}}, and {{Understanding}} in the {{Age}} of {{Data}}},
  shorttitle = {Climbing towards {{NLU}}},
  booktitle = {Proceedings of the 58th {{Annual Meeting}} of the {{Association}} for {{Computational Linguistics}}},
  author = {Bender, Emily M. and Koller, Alexander},
  year = {2020},
  month = jul,
  pages = {5185--5198},
  publisher = {{Association for Computational Linguistics}},
  address = {{Online}},
  doi = {10.18653/v1/2020.acl-main.463},
  urldate = {2021-10-14}
}

@inproceedings{benderDangersStochasticParrots2021,
  title = {On the {{Dangers}} of {{Stochastic Parrots}}: {{Can Language Models Be Too Big}}? \&\#x1f99c;},
  shorttitle = {On the {{Dangers}} of {{Stochastic Parrots}}},
  booktitle = {Proceedings of the 2021 {{ACM Conference}} on {{Fairness}}, {{Accountability}}, and {{Transparency}}},
  author = {Bender, Emily M. and Gebru, Timnit and {McMillan-Major}, Angelina and Shmitchell, Shmargaret},
  year = {2021},
  month = mar,
  series = {{{FAccT}} '21},
  pages = {610--623},
  publisher = {{Association for Computing Machinery}},
  address = {{New York, NY, USA}},
  doi = {10.1145/3442188.3445922},
  urldate = {2021-04-21},
  isbn = {978-1-4503-8309-7}
}

@inproceedings{bengioNeuralProbabilisticLanguage2000,
  title = {A {{Neural Probabilistic Language Model}}},
  booktitle = {Advances in {{Neural Information Processing Systems}}},
  author = {Bengio, Yoshua and Ducharme, R{\'e}jean and Vincent, Pascal},
  year = {2000},
  volume = {13},
  publisher = {{MIT Press}},
  urldate = {2023-08-04}
}

@article{betker2023improving,
  title = {Improving Image Generation with Better Captions},
  author = {Betker, James and Goh, Gabriel and Jing, Li and Brooks, Tim and Wang, Jianfeng and Li, Linjie and Ouyang, Long and Zhuang, Juntang and Lee, Joyce and Guo, Yufei and others},
  year = {2023},
  journal = {Computer Science. https://cdn. openai. com/papers/dall-e-3. pdf}
}

@article{blockAdvertisementSemanticsPsychology1986,
  title = {Advertisement for a {{Semantics}} for {{Psychology}}},
  author = {Block, Ned},
  year = {1986},
  month = apr,
  journal = {Midwest Studies in Philosophy},
  volume = {10},
  pages = {615--678},
  doi = {10.1111/j.1475-4975.1987.tb00558.x},
  urldate = {2023-03-30},
  langid = {english}
}

@article{blockPsychologismBehaviorism1981,
  title = {Psychologism and {{Behaviorism}}},
  author = {Block, Ned},
  year = {1981},
  journal = {The Philosophical Review},
  volume = {90},
  number = {1},
  eprint = {2184371},
  eprinttype = {jstor},
  pages = {5--43},
  publisher = {{[Duke University Press, Philosophical Review]}},
  issn = {0031-8108},
  doi = {10.2307/2184371},
  urldate = {2023-10-17}
}

@article{boledaDistributionalSemanticsLinguistic2020,
  title = {Distributional {{Semantics}} and {{Linguistic Theory}}},
  author = {Boleda, Gemma},
  year = {2020},
  journal = {Annual Review of Linguistics},
  volume = {6},
  number = {1},
  pages = {213--234},
  doi = {10.1146/annurev-linguistics-011619-030303},
  urldate = {2022-11-11}
}

@article{brownLanguageModelsAre2020,
  title = {Language {{Models}} Are {{Few-Shot Learners}}},
  author = {Brown, Tom B. and Mann, Benjamin and Ryder, Nick and Subbiah, Melanie and Kaplan, Jared and Dhariwal, Prafulla and Neelakantan, Arvind and Shyam, Pranav and Sastry, Girish and Askell, Amanda and Agarwal, Sandhini and {Herbert-Voss}, Ariel and Krueger, Gretchen and Henighan, Tom and Child, Rewon and Ramesh, Aditya and Ziegler, Daniel M. and Wu, Jeffrey and Winter, Clemens and Hesse, Christopher and Chen, Mark and Sigler, Eric and Litwin, Mateusz and Gray, Scott and Chess, Benjamin and Clark, Jack and Berner, Christopher and McCandlish, Sam and Radford, Alec and Sutskever, Ilya and Amodei, Dario},
  year = {2020},
  month = jul,
  journal = {arXiv:2005.14165 [cs]},
  eprint = {2005.14165},
  primaryclass = {cs},
  urldate = {2020-11-10},
  archiveprefix = {arxiv}
}

@misc{bubeckSparksArtificialGeneral2023,
  title = {Sparks of {{Artificial General Intelligence}}: {{Early}} Experiments with {{GPT-4}}},
  shorttitle = {Sparks of {{Artificial General Intelligence}}},
  author = {Bubeck, S{\'e}bastien and Chandrasekaran, Varun and Eldan, Ronen and Gehrke, Johannes and Horvitz, Eric and Kamar, Ece and Lee, Peter and Lee, Yin Tat and Li, Yuanzhi and Lundberg, Scott and Nori, Harsha and Palangi, Hamid and Ribeiro, Marco Tulio and Zhang, Yi},
  year = {2023},
  month = mar,
  number = {arXiv:2303.12712},
  eprint = {2303.12712},
  primaryclass = {cs},
  publisher = {{arXiv}},
  doi = {10.48550/arXiv.2303.12712},
  urldate = {2023-03-28},
  archiveprefix = {arxiv}
}

@article{bucknerBlackBoxesUnflattering2021,
  title = {Black {{Boxes}} or {{Unflattering Mirrors}}? {{Comparative Bias}} in the {{Science}} of {{Machine Behaviour}}},
  shorttitle = {Black {{Boxes}} or {{Unflattering Mirrors}}?},
  author = {Buckner, Cameron},
  year = {2021},
  month = apr,
  journal = {The British Journal for the Philosophy of Science},
  pages = {000--000},
  publisher = {{The University of Chicago Press}},
  issn = {0007-0882},
  doi = {10.1086/714960},
  urldate = {2023-08-09}
}

@book{bucknerDeepLearningRational2023,
  title = {From {{Deep Learning}} to {{Rational Machines}}: {{What}} the {{History}} of {{Philosophy Can Teach Us}} about the {{Future}} of {{Artificial Intelligence}}},
  shorttitle = {From {{Deep Learning}} to {{Rational Machines}}},
  author = {Buckner, Cameron J.},
  year = {2023},
  month = dec,
  publisher = {{Oxford University Press}},
  address = {{Oxford, New York}},
  isbn = {978-0-19-765330-2}
}

@incollection{bucknerUnderstandingAssociativeCognitive2017,
  title = {Understanding {{Associative}} and {{Cognitive Explanations}} in {{Comparative Psychology}}},
  booktitle = {The {{Routledge Handbook}} of {{Philosophy}} of {{Animal Minds}}},
  author = {Buckner, Cameron},
  year = {2017},
  publisher = {{Routledge}},
  isbn = {978-1-315-74225-0}
}

@article{butlinSharingOurConcepts2021,
  title = {Sharing {{Our Concepts}} with {{Machines}}},
  author = {Butlin, Patrick},
  year = {2021},
  month = nov,
  journal = {Erkenntnis},
  issn = {1572-8420},
  doi = {10.1007/s10670-021-00491-w},
  urldate = {2022-11-11},
  langid = {english}
}

@book{carnieSyntaxGenerativeIntroduction2021,
  title = {Syntax: {{A Generative Introduction}}},
  shorttitle = {Syntax},
  author = {Carnie, Andrew},
  year = {2021},
  month = apr,
  publisher = {{John Wiley \& Sons}},
  googlebooks = {sh0gEAAAQBAJ},
  isbn = {978-1-119-56923-7},
  langid = {english}
}

@misc{choLearningPhraseRepresentations2014,
  title = {Learning {{Phrase Representations}} Using {{RNN Encoder-Decoder}} for {{Statistical Machine Translation}}},
  author = {Cho, Kyunghyun and {van Merrienboer}, Bart and Gulcehre, Caglar and Bahdanau, Dzmitry and Bougares, Fethi and Schwenk, Holger and Bengio, Yoshua},
  year = {2014},
  month = sep,
  number = {arXiv:1406.1078},
  eprint = {1406.1078},
  primaryclass = {cs, stat},
  publisher = {{arXiv}},
  doi = {10.48550/arXiv.1406.1078},
  urldate = {2023-10-17},
  archiveprefix = {arxiv}
}

@misc{cholletMeasureIntelligence2019,
  title = {On the {{Measure}} of {{Intelligence}}},
  author = {Chollet, Fran{\c c}ois},
  year = {2019},
  month = nov,
  number = {arXiv:1911.01547},
  eprint = {1911.01547},
  primaryclass = {cs},
  publisher = {{arXiv}},
  doi = {10.48550/arXiv.1911.01547},
  urldate = {2023-12-19},
  archiveprefix = {arxiv}
}

@incollection{chomskyKnowledqeLanquaqeIts2000,
  title = {Knowledqe of {{Lanquaqe}}: {{Its Nature}}, {{Oriqin}} and {{Use}}},
  shorttitle = {Knowledqe of {{Lanquaqe}}},
  booktitle = {Perspectives in the {{Philosophy}} of {{Language}}: {{A Concise Anthology}}},
  author = {Chomsky, Noam},
  editor = {Stainton, Robert J.},
  year = {2000},
  pages = {3},
  publisher = {{Broadview Press}}
}

@book{chomskySyntacticStructures1957,
  title = {Syntactic {{Structures}}},
  author = {Chomsky, Noam},
  year = {1957},
  publisher = {{Mouton}}
}

@inproceedings{christianoDeepReinforcementLearning2017,
  title = {Deep {{Reinforcement Learning}} from {{Human Preferences}}},
  booktitle = {Advances in {{Neural Information Processing Systems}}},
  author = {Christiano, Paul F and Leike, Jan and Brown, Tom and Martic, Miljan and Legg, Shane and Amodei, Dario},
  year = {2017},
  volume = {30},
  publisher = {{Curran Associates, Inc.}},
  urldate = {2023-10-17}
}

@inproceedings{conklinMetaLearningCompositionallyGeneralize2021,
  title = {Meta-{{Learning}} to {{Compositionally Generalize}}},
  booktitle = {Proceedings of the 59th {{Annual Meeting}} of the {{Association}} for {{Computational Linguistics}} and the 11th {{International Joint Conference}} on {{Natural Language Processing}} ({{Volume}} 1: {{Long Papers}})},
  author = {Conklin, Henry and Wang, Bailin and Smith, Kenny and Titov, Ivan},
  year = {2021},
  month = aug,
  pages = {3322--3335},
  publisher = {{Association for Computational Linguistics}},
  address = {{Online}},
  doi = {10.18653/v1/2021.acl-long.258},
  urldate = {2023-08-21}
}

@inproceedings{csordasCTLEvaluatingGeneralization2022,
  title = {{{CTL}}++: {{Evaluating Generalization}} on {{Never-Seen Compositional Patterns}} of {{Known Functions}}, and {{Compatibility}} of {{Neural Representations}}},
  shorttitle = {{{CTL}}++},
  booktitle = {Proceedings of the 2022 {{Conference}} on {{Empirical Methods}} in {{Natural Language Processing}}},
  author = {Csord{\'a}s, R{\'o}bert and Irie, Kazuki and Schmidhuber, Juergen},
  year = {2022},
  month = dec,
  pages = {9758--9767},
  publisher = {{Association for Computational Linguistics}},
  address = {{Abu Dhabi, United Arab Emirates}},
  doi = {10.18653/v1/2022.emnlp-main.662},
  urldate = {2023-10-03}
}

@article{dabrowskaWhatExactlyUniversal2015,
  title = {What Exactly Is {{Universal Grammar}}, and Has Anyone Seen It?},
  author = {D{\k{a}}browska, Ewa},
  year = {2015},
  journal = {Frontiers in Psychology},
  volume = {6},
  issn = {1664-1078},
  urldate = {2023-10-18}
}

@article{firthSynopsisLinguisticTheory1957,
  title = {A Synopsis of Linguistic Theory, 1930-1955},
  author = {Firth, John R.},
  year = {1957},
  journal = {Studies in linguistic analysis},
  publisher = {{Basil Blackwell}}
}

@article{fodorConnectionismCognitiveArchitecture1988,
  title = {Connectionism and Cognitive Architecture: {{A}} Critical Analysis},
  shorttitle = {Connectionism and Cognitive Architecture},
  author = {Fodor, Jerry A. and Pylyshyn, Zenon W.},
  year = {1988},
  month = mar,
  journal = {Cognition},
  volume = {28},
  number = {1},
  pages = {3--71},
  issn = {0010-0277},
  doi = {10.1016/0010-0277(88)90031-5},
  urldate = {2023-10-05}
}

@book{fodorLanguageThought1975,
  title = {The {{Language}} of {{Thought}}},
  author = {Fodor, Jerry A.},
  year = {1975},
  publisher = {{Harvard University Press}}
}

@article{grandSemanticProjectionRecovers2022,
  title = {Semantic Projection Recovers Rich Human Knowledge of Multiple Object Features from Word Embeddings},
  author = {Grand, Gabriel and Blank, Idan Asher and Pereira, Francisco and Fedorenko, Evelina},
  year = {2022},
  month = jul,
  journal = {Nature Human Behaviour},
  volume = {6},
  number = {7},
  pages = {975--987},
  publisher = {{Nature Publishing Group}},
  issn = {2397-3374},
  doi = {10.1038/s41562-022-01316-8},
  urldate = {2023-09-25},
  copyright = {2022 The Author(s), under exclusive licence to Springer Nature Limited},
  langid = {english}
}

@article{grynbaumTimesSuesOpenAI2023,
  title = {The {{Times Sues OpenAI}} and {{Microsoft Over A}}.{{I}}. {{Use}} of {{Copyrighted Work}}},
  author = {Grynbaum, Michael M. and Mac, Ryan},
  year = {2023},
  month = dec,
  journal = {The New York Times},
  issn = {0362-4331},
  urldate = {2023-12-31},
  chapter = {Business},
  langid = {american}
}

@article{harnadSymbolGroundingProblem1990,
  title = {The Symbol Grounding Problem},
  author = {Harnad, Stevan},
  year = {1990},
  month = jun,
  journal = {Physica D: Nonlinear Phenomena},
  volume = {42},
  number = {1},
  pages = {335--346},
  issn = {0167-2789},
  doi = {10.1016/0167-2789(90)90087-6},
  urldate = {2021-04-20},
  langid = {english}
}

@article{harrisDistributionalStructure1954,
  title = {Distributional Structure},
  author = {Harris, Zellig S.},
  year = {1954},
  journal = {Word},
  volume = {10},
  pages = {146--162},
  doi = {10.1080/00437956.1954.11659520}
}

@article{haWorldModels2018,
  title = {World {{Models}}},
  author = {Ha, David and Schmidhuber, J{\"u}rgen},
  year = {2018},
  month = mar,
  eprint = {1803.10122},
  primaryclass = {cs, stat},
  doi = {10.5281/zenodo.1207631},
  urldate = {2023-12-11},
  archiveprefix = {arxiv}
}

@inproceedings{heLargeLanguageModels2023,
  title = {Large {{Language Models}} as {{Zero-Shot Conversational Recommenders}}},
  booktitle = {Proceedings of the 32nd {{ACM International Conference}} on {{Information}} and {{Knowledge Management}}},
  author = {He, Zhankui and Xie, Zhouhang and Jha, Rahul and Steck, Harald and Liang, Dawen and Feng, Yesu and Majumder, Bodhisattwa Prasad and Kallus, Nathan and Mcauley, Julian},
  year = {2023},
  month = oct,
  series = {{{CIKM}} '23},
  pages = {720--730},
  publisher = {{Association for Computing Machinery}},
  address = {{New York, NY, USA}},
  doi = {10.1145/3583780.3614949},
  urldate = {2024-01-02},
  isbn = {9798400701245}
}

@article{herboldLargescaleComparisonHumanwritten2023,
  title = {A Large-Scale Comparison of Human-Written versus {{ChatGPT-generated}} Essays},
  author = {Herbold, Steffen and {Hautli-Janisz}, Annette and Heuer, Ute and Kikteva, Zlata and Trautsch, Alexander},
  year = {2023},
  month = oct,
  journal = {Scientific Reports},
  volume = {13},
  number = {1},
  pages = {18617},
  publisher = {{Nature Publishing Group}},
  issn = {2045-2322},
  doi = {10.1038/s41598-023-45644-9},
  urldate = {2023-12-19},
  copyright = {2023 The Author(s)},
  langid = {english}
}

@article{hochreiterLongShortTermMemory1997,
  title = {Long {{Short-Term Memory}}},
  author = {Hochreiter, Sepp and Schmidhuber, J{\"u}rgen},
  year = {1997},
  month = nov,
  journal = {Neural Computation},
  volume = {9},
  number = {8},
  pages = {1735--1780},
  issn = {0899-7667},
  doi = {10.1162/neco.1997.9.8.1735},
  urldate = {2023-10-17}
}

@inproceedings{huebnerBabyBERTaLearningMore2021,
  title = {{{BabyBERTa}}: {{Learning More Grammar With Small-Scale Child-Directed Language}}},
  shorttitle = {{{BabyBERTa}}},
  booktitle = {Proceedings of the 25th {{Conference}} on {{Computational Natural Language Learning}}},
  author = {Huebner, Philip A. and Sulem, Elior and Cynthia, Fisher and Roth, Dan},
  editor = {Bisazza, Arianna and Abend, Omri},
  year = {2021},
  month = nov,
  pages = {624--646},
  publisher = {{Association for Computational Linguistics}},
  address = {{Online}},
  doi = {10.18653/v1/2021.conll-1.49},
  urldate = {2023-12-19}
}

@book{humeTreatiseHumanNature1978,
  title = {A {{Treatise}} of {{Human Nature}}},
  author = {Hume, David},
  editor = {{Selby-Bigge}, L. A. and Nidditch, P. H.},
  year = {1978},
  month = nov,
  edition = {2nd edition},
  publisher = {{Oxford University Press}},
  address = {{Oxford}},
  isbn = {978-0-19-824588-9},
  langid = {english}
}

@article{Hupkes2023,
  title = {A Taxonomy and Review of Generalization Research in {{NLP}}},
  author = {Hupkes, Dieuwke and Giulianelli, Mario and Dankers, Verna and Artetxe, Mikel and Elazar, Yanai and Pimentel, Tiago and Christodoulopoulos, Christos and Lasri, Karim and Saphra, Naomi and Sinclair, Arabella and Ulmer, Dennis and Schottmann, Florian and Batsuren, Khuyagbaatar and Sun, Kaiser and Sinha, Koustuv and Khalatbari, Leila and Ryskina, Maria and Frieske, Rita and Cotterell, Ryan and Jin, Zhijing},
  year = {2023},
  month = oct,
  journal = {Nature Machine Intelligence},
  volume = {5},
  number = {10},
  pages = {1161--1174},
  issn = {2522-5839},
  doi = {10.1038/s42256-023-00729-y}
}

@book{jelinekStatisticalMethodsSpeech1998,
  title = {Statistical Methods for Speech Recognition},
  author = {Jelinek, Frederick},
  year = {1998},
  month = jan,
  publisher = {{MIT Press}},
  address = {{Cambridge, MA, USA}},
  isbn = {978-0-262-10066-3}
}

@misc{jonesDoesGPT4Pass2023,
  title = {Does {{GPT-4 Pass}} the {{Turing Test}}?},
  author = {Jones, Cameron and Bergen, Benjamin},
  year = {2023},
  month = oct,
  number = {arXiv:2310.20216},
  eprint = {2310.20216},
  primaryclass = {cs},
  publisher = {{arXiv}},
  doi = {10.48550/arXiv.2310.20216},
  urldate = {2023-12-19},
  archiveprefix = {arxiv}
}

@misc{karhadeGPT4ModelsOne2023,
  title = {{{GPT-4}}: 8 {{Models}} in {{One}} ; {{The Secret}} Is {{Out}}},
  shorttitle = {{{GPT-4}}},
  author = {Karhade, Mandar},
  year = {2023},
  month = jul,
  journal = {Medium},
  urldate = {2023-12-19},
  langid = {english}
}

@article{kasneciChatGPTGoodOpportunities2023,
  title = {{{ChatGPT}} for Good? {{On}} Opportunities and Challenges of Large Language Models for Education},
  shorttitle = {{{ChatGPT}} for Good?},
  author = {Kasneci, Enkelejda and Sessler, Kathrin and K{\"u}chemann, Stefan and Bannert, Maria and Dementieva, Daryna and Fischer, Frank and Gasser, Urs and Groh, Georg and G{\"u}nnemann, Stephan and H{\"u}llermeier, Eyke and Krusche, Stephan and Kutyniok, Gitta and Michaeli, Tilman and Nerdel, Claudia and Pfeffer, J{\"u}rgen and Poquet, Oleksandra and Sailer, Michael and Schmidt, Albrecht and Seidel, Tina and Stadler, Matthias and Weller, Jochen and Kuhn, Jochen and Kasneci, Gjergji},
  year = {2023},
  month = apr,
  journal = {Learning and Individual Differences},
  volume = {103},
  pages = {102274},
  issn = {1041-6080},
  doi = {10.1016/j.lindif.2023.102274},
  urldate = {2024-01-03}
}

@inproceedings{keysersMeasuringCompositionalGeneralization2019,
  title = {Measuring {{Compositional Generalization}}: {{A Comprehensive Method}} on {{Realistic Data}}},
  shorttitle = {Measuring {{Compositional Generalization}}},
  booktitle = {International {{Conference}} on {{Learning Representations}}},
  author = {Keysers, Daniel and Sch{\"a}rli, Nathanael and Scales, Nathan and Buisman, Hylke and Furrer, Daniel and Kashubin, Sergii and Momchev, Nikola and Sinopalnikov, Danila and Stafiniak, Lukasz and Tihon, Tibor and Tsarkov, Dmitry and Wang, Xiao and van Zee, Marc and Bousquet, Olivier},
  year = {2019},
  month = sep,
  urldate = {2023-08-21},
  langid = {english}
}

@misc{kheiriSentimentGPTExploitingGPT2023,
  title = {{{SentimentGPT}}: {{Exploiting GPT}} for {{Advanced Sentiment Analysis}} and Its {{Departure}} from {{Current Machine Learning}}},
  shorttitle = {{{SentimentGPT}}},
  author = {Kheiri, Kiana and Karimi, Hamid},
  year = {2023},
  month = jul,
  number = {arXiv:2307.10234},
  eprint = {2307.10234},
  primaryclass = {cs},
  publisher = {{arXiv}},
  doi = {10.48550/arXiv.2307.10234},
  urldate = {2024-01-03},
  archiveprefix = {arxiv}
}

@inproceedings{kimCOGSCompositionalGeneralization2020,
  title = {{{COGS}}: {{A Compositional Generalization Challenge Based}} on {{Semantic Interpretation}}},
  shorttitle = {{{COGS}}},
  booktitle = {Proceedings of the 2020 {{Conference}} on {{Empirical Methods}} in {{Natural Language Processing}} ({{EMNLP}})},
  author = {Kim, Najoung and Linzen, Tal},
  year = {2020},
  month = nov,
  pages = {9087--9105},
  publisher = {{Association for Computational Linguistics}},
  address = {{Online}},
  doi = {10.18653/v1/2020.emnlp-main.731},
  urldate = {2023-08-21}
}

@book{kripkeNamingNecessity1980,
  title = {Naming and {{Necessity}}},
  author = {Kripke, Saul},
  year = {1980},
  publisher = {{Harvard University Press}},
  address = {{Cambridge, MA}}
}

@article{lakeBuildingMachinesThat2017,
  title = {Building Machines That Learn and Think like People},
  author = {Lake, Brenden M. and Ullman, Tomer D. and Tenenbaum, Joshua B. and Gershman, Samuel J.},
  year = {2017},
  journal = {Behavioral and Brain Sciences},
  volume = {40},
  publisher = {{Cambridge University Press}},
  issn = {0140-525X, 1469-1825},
  doi = {10.1017/S0140525X16001837},
  urldate = {2021-03-15},
  langid = {english}
}

@inproceedings{lakeGeneralizationSystematicityCompositional2018,
  title = {Generalization without {{Systematicity}}: {{On}} the {{Compositional Skills}} of {{Sequence-to-Sequence Recurrent Networks}}},
  shorttitle = {Generalization without {{Systematicity}}},
  booktitle = {Proceedings of the 35th {{International Conference}} on {{Machine Learning}}},
  author = {Lake, Brenden and Baroni, Marco},
  year = {2018},
  month = jul,
  pages = {2873--2882},
  publisher = {{PMLR}},
  issn = {2640-3498},
  urldate = {2023-08-21},
  langid = {english}
}

@article{lakeHumanlikeSystematicGeneralization2023,
  title = {Human-like Systematic Generalization through a Meta-Learning Neural Network},
  author = {Lake, Brenden M. and Baroni, Marco},
  year = {2023},
  month = oct,
  journal = {Nature},
  pages = {1--7},
  publisher = {{Nature Publishing Group}},
  issn = {1476-4687},
  doi = {10.1038/s41586-023-06668-3},
  urldate = {2023-10-30},
  copyright = {2023 The Author(s)},
  langid = {english}
}

@article{lasnikGovernmentBindingPrinciples2010,
  title = {Government{\textendash}Binding/Principles and Parameters Theory},
  author = {Lasnik, Howard and Lohndal, Terje},
  year = {2010},
  journal = {WIREs Cognitive Science},
  volume = {1},
  number = {1},
  pages = {40--50},
  issn = {1939-5086},
  doi = {10.1002/wcs.35},
  urldate = {2023-10-18},
  copyright = {Copyright {\textcopyright} 2009 John Wiley \& Sons, Ltd.},
  langid = {english}
}

@misc{lavechinBabySLMLanguageacquisitionfriendlyBenchmark2023,
  title = {{{BabySLM}}: Language-Acquisition-Friendly Benchmark of Self-Supervised Spoken Language Models},
  shorttitle = {{{BabySLM}}},
  author = {Lavechin, Marvin and Sy, Yaya and Titeux, Hadrien and Bland{\'o}n, Mar{\'i}a Andrea Cruz and R{\"a}s{\"a}nen, Okko and Bredin, Herv{\'e} and Dupoux, Emmanuel and Cristia, Alejandrina},
  year = {2023},
  month = jun,
  number = {arXiv:2306.01506},
  eprint = {2306.01506},
  primaryclass = {cs, eess, stat},
  publisher = {{arXiv}},
  doi = {10.48550/arXiv.2306.01506},
  urldate = {2023-08-18},
  archiveprefix = {arxiv}
}

@misc{lecunPathAutonomousMachine,
  title = {A {{Path Towards Autonomous Machine Intelligence}}},
  author = {LeCun, Yann},
  urldate = {2023-12-11},
  langid = {english}
}

@inproceedings{leeTeachingArithmeticSmall2023,
  title = {Teaching {{Arithmetic}} to {{Small Transformers}}},
  booktitle = {The 3rd {{Workshop}} on {{Mathematical Reasoning}} and {{AI}} at {{NeurIPS}}'23},
  author = {Lee, Nayoung and Sreenivasan, Kartik and Lee, Jason and Lee, Kangwook and Papailiopoulos, Dimitris},
  year = {2023},
  month = oct,
  urldate = {2023-12-14},
  langid = {english}
}

@misc{lewkowyczSolvingQuantitativeReasoning2022,
  title = {Solving {{Quantitative Reasoning Problems}} with {{Language Models}}},
  author = {Lewkowycz, Aitor and Andreassen, Anders and Dohan, David and Dyer, Ethan and Michalewski, Henryk and Ramasesh, Vinay and Slone, Ambrose and Anil, Cem and Schlag, Imanol and {Gutman-Solo}, Theo and Wu, Yuhuai and Neyshabur, Behnam and {Gur-Ari}, Guy and Misra, Vedant},
  year = {2022},
  month = jun,
  number = {arXiv:2206.14858},
  eprint = {2206.14858},
  primaryclass = {cs},
  publisher = {{arXiv}},
  doi = {10.48550/arXiv.2206.14858},
  urldate = {2024-01-03},
  archiveprefix = {arxiv}
}

@misc{liangCanLargeLanguage2023,
  title = {Can Large Language Models Provide Useful Feedback on Research Papers? {{A}} Large-Scale Empirical Analysis},
  shorttitle = {Can Large Language Models Provide Useful Feedback on Research Papers?},
  author = {Liang, Weixin and Zhang, Yuhui and Cao, Hancheng and Wang, Binglu and Ding, Daisy and Yang, Xinyu and Vodrahalli, Kailas and He, Siyu and Smith, Daniel and Yin, Yian and McFarland, Daniel and Zou, James},
  year = {2023},
  month = oct,
  number = {arXiv:2310.01783},
  eprint = {2310.01783},
  primaryclass = {cs},
  publisher = {{arXiv}},
  doi = {10.48550/arXiv.2310.01783},
  urldate = {2024-01-03},
  archiveprefix = {arxiv}
}

@article{longBabyViewCameraDesigning2023,
  title = {The {{BabyView}} Camera: {{Designing}} a New Head-Mounted Camera to Capture Children's Early Social and Visual Environments},
  shorttitle = {The {{BabyView}} Camera},
  author = {Long, Bria and Goodin, Sarah and Kachergis, George and Marchman, Virginia A. and Radwan, Samaher F. and Sparks, Robert Z. and Xiang, Violet and Zhuang, Chengxu and Hsu, Oliver and Newman, Brett and Yamins, Daniel L. K. and Frank, Michael C.},
  year = {2023},
  month = sep,
  journal = {Behavior Research Methods},
  issn = {1554-3528},
  doi = {10.3758/s13428-023-02206-1},
  urldate = {2023-12-19},
  langid = {english}
}

@incollection{macdonaldClassicismVsConnectionism1995,
  title = {Classicism {{Vs}}. {{Connectionism}}},
  booktitle = {Connectionism: {{Debates}} on {{Psychological Explanation}}},
  author = {Macdonald, Cynthia},
  editor = {Macdonald, Cynthia and Macdonald, Graham F.},
  year = {1995},
  publisher = {{Blackwell}}
}

@misc{mandelkernLanguageModelsRefer2023,
  title = {Do {{Language Models Refer}}?},
  author = {Mandelkern, Matthew and Linzen, Tal},
  year = {2023},
  month = aug,
  number = {arXiv:2308.05576},
  eprint = {2308.05576},
  primaryclass = {cs},
  publisher = {{arXiv}},
  doi = {10.48550/arXiv.2308.05576},
  urldate = {2023-09-22},
  archiveprefix = {arxiv}
}

@book{marconiLexicalCompetence1997,
  title = {Lexical {{Competence}}},
  author = {Marconi, Diego},
  year = {1997},
  publisher = {{MIT Press}},
  googlebooks = {lcrEq\_7o5m0C},
  isbn = {978-0-262-13333-3},
  langid = {english}
}

@misc{mccoyEmbersAutoregressionUnderstanding2023,
  title = {Embers of {{Autoregression}}: {{Understanding Large Language Models Through}} the {{Problem They}} Are {{Trained}} to {{Solve}}},
  shorttitle = {Embers of {{Autoregression}}},
  author = {McCoy, R. Thomas and Yao, Shunyu and Friedman, Dan and Hardy, Matthew and Griffiths, Thomas L.},
  year = {2023},
  month = sep,
  number = {arXiv:2309.13638},
  eprint = {2309.13638},
  primaryclass = {cs},
  publisher = {{arXiv}},
  doi = {10.48550/arXiv.2309.13638},
  urldate = {2023-09-27},
  archiveprefix = {arxiv}
}

@misc{mcgrathPropertiesLoTsFootprints2023,
  title = {Properties of {{LoTs}}: {{The}} Footprints or the Bear Itself?},
  shorttitle = {Properties of {{LoTs}}},
  author = {McGrath, Sam and Russin, Jacob and Pavlick, Ellie and Feiman, Roman},
  year = {2023},
  month = apr,
  publisher = {{PsyArXiv}},
  doi = {10.31234/osf.io/tdw34},
  urldate = {2023-08-21},
  langid = {american}
}

@article{mikolovEfficientEstimationWord2013,
  title = {Efficient {{Estimation}} of {{Word Representations}} in {{Vector Space}}},
  author = {Mikolov, Tomas and Chen, Kai and Corrado, Greg and Dean, Jeffrey},
  year = {2013},
  month = sep,
  journal = {arXiv:1301.3781 [cs]},
  eprint = {1301.3781},
  primaryclass = {cs},
  urldate = {2020-01-27},
  archiveprefix = {arxiv}
}

@incollection{milliereLanguageModelsModelsforthcoming,
  title = {Language {{Models}} as {{Models}} of {{Language}}},
  booktitle = {The {{Oxford Handbook}} of the {{Philosophy}} of {{Linguistics}}},
  author = {Milli{\`e}re, Rapha{\"e}l},
  editor = {Nefdt, Ryan and Dupre, Gabe and Jain, Kate Hazel},
  year = {forthcoming},
  publisher = {{Oxford University Press}},
  address = {{Oxford}}
}

@misc{mirchandaniLargeLanguageModels2023,
  title = {Large {{Language Models}} as {{General Pattern Machines}}},
  author = {Mirchandani, Suvir and Xia, Fei and Florence, Pete and Ichter, Brian and Driess, Danny and Arenas, Montserrat Gonzalez and Rao, Kanishka and Sadigh, Dorsa and Zeng, Andy},
  year = {2023},
  month = jul,
  number = {arXiv:2307.04721},
  eprint = {2307.04721},
  primaryclass = {cs},
  publisher = {{arXiv}},
  doi = {10.48550/arXiv.2307.04721},
  urldate = {2023-07-27},
  archiveprefix = {arxiv}
}

@inproceedings{mirowskiCoWritingScreenplaysTheatre2023,
  title = {Co-{{Writing Screenplays}} and {{Theatre Scripts}} with {{Language Models}}: {{Evaluation}} by {{Industry Professionals}}},
  shorttitle = {Co-{{Writing Screenplays}} and {{Theatre Scripts}} with {{Language Models}}},
  booktitle = {Proceedings of the 2023 {{CHI Conference}} on {{Human Factors}} in {{Computing Systems}}},
  author = {Mirowski, Piotr and Mathewson, Kory W. and Pittman, Jaylen and Evans, Richard},
  year = {2023},
  month = apr,
  series = {{{CHI}} '23},
  pages = {1--34},
  publisher = {{Association for Computing Machinery}},
  address = {{New York, NY, USA}},
  doi = {10.1145/3544548.3581225},
  urldate = {2024-01-02},
  isbn = {978-1-4503-9421-5}
}

@misc{molloVectorGroundingProblem2023,
  title = {The {{Vector Grounding Problem}}},
  author = {Mollo, Dimitri Coelho and Milli{\`e}re, Rapha{\"e}l},
  year = {2023},
  month = apr,
  number = {arXiv:2304.01481},
  eprint = {2304.01481},
  primaryclass = {cs},
  publisher = {{arXiv}},
  doi = {10.48550/arXiv.2304.01481},
  urldate = {2023-08-04},
  archiveprefix = {arxiv}
}

@misc{murtyGrokkingHierarchicalStructure2023,
  title = {Grokking of {{Hierarchical Structure}} in {{Vanilla Transformers}}},
  author = {Murty, Shikhar and Sharma, Pratyusha and Andreas, Jacob and Manning, Christopher D.},
  year = {2023},
  month = may,
  number = {arXiv:2305.18741},
  eprint = {2305.18741},
  primaryclass = {cs},
  publisher = {{arXiv}},
  doi = {10.48550/arXiv.2305.18741},
  urldate = {2023-08-01},
  archiveprefix = {arxiv}
}

@inproceedings{ontanonMakingTransformersSolve2022,
  title = {Making {{Transformers Solve Compositional Tasks}}},
  booktitle = {Proceedings of the 60th {{Annual Meeting}} of the {{Association}} for {{Computational Linguistics}} ({{Volume}} 1: {{Long Papers}})},
  author = {Ontanon, Santiago and Ainslie, Joshua and Fisher, Zachary and Cvicek, Vaclav},
  year = {2022},
  month = may,
  pages = {3591--3607},
  publisher = {{Association for Computational Linguistics}},
  address = {{Dublin, Ireland}},
  doi = {10.18653/v1/2022.acl-long.251},
  urldate = {2022-07-08}
}

@misc{openaiGPT4TechnicalReport2023,
  title = {{{GPT-4 Technical Report}}},
  author = {OpenAI},
  year = {2023},
  month = mar,
  number = {arXiv:2303.08774},
  eprint = {2303.08774},
  primaryclass = {cs},
  publisher = {{arXiv}},
  doi = {10.48550/arXiv.2303.08774},
  urldate = {2023-03-28},
  archiveprefix = {arxiv}
}

@misc{openaiGPT4VIsionSystem2023,
  title = {{{GPT-4V}}(Ision) {{System Card}}},
  author = {{OpenAI}},
  year = {2023},
  month = sep
}

@misc{openaiIntroducingChatGPT2022,
  title = {Introducing {{ChatGPT}}},
  author = {{OpenAI}},
  year = {2022},
  month = nov,
  journal = {OpenAI Blog},
  urldate = {2023-10-29},
  langid = {american}
}

@article{osgoodNatureMeasurementMeaning1952,
  title = {The Nature and Measurement of Meaning},
  author = {Osgood, Charles E.},
  year = {1952},
  month = may,
  journal = {Psychological bulletin},
  volume = {49},
  number = {3},
  pages = {197--237},
  issn = {1939-1455},
  doi = {10.1037/h0055737},
  urldate = {2023-08-03},
  langid = {english},
  pmid = {14930159}
}

@article{pavlickSymbolsGroundingLarge2023,
  title = {Symbols and Grounding in Large Language Models},
  author = {Pavlick, Ellie},
  year = {2023},
  month = jun,
  journal = {Philosophical Transactions of the Royal Society A: Mathematical, Physical and Engineering Sciences},
  volume = {381},
  number = {2251},
  pages = {20220041},
  publisher = {{Royal Society}},
  doi = {10.1098/rsta.2022.0041},
  urldate = {2023-08-05}
}

@article{pearlPovertyStimulusTears2022,
  title = {Poverty of the {{Stimulus Without Tears}}},
  author = {Pearl, Lisa},
  year = {2022},
  month = oct,
  journal = {Language Learning and Development},
  volume = {18},
  number = {4},
  pages = {415--454},
  publisher = {{Routledge}},
  issn = {1547-5441},
  doi = {10.1080/15475441.2021.1981908},
  urldate = {2023-07-30}
}

@misc{piantadosiMeaningReferenceLarge2022,
  title = {Meaning without Reference in Large Language Models},
  author = {Piantadosi, Steven and Hill, Felix},
  year = {2022},
  month = aug,
  number = {arXiv:2208.02957},
  eprint = {2208.02957},
  primaryclass = {cs},
  publisher = {{arXiv}},
  doi = {10.48550/arXiv.2208.02957},
  urldate = {2022-11-10},
  archiveprefix = {arxiv}
}

@misc{piantadosiModernLanguageModels2023,
  title = {Modern Language Models Refute {{Chomsky}}'s Approach to Language},
  author = {Piantadosi, Steven},
  year = {2023},
  month = mar,
  publisher = {{LingBuzz}},
  urldate = {2023-07-30},
  archiveprefix = {LingBuzz}
}

@article{pinkerLanguageConnectionismAnalysis1988,
  title = {On Language and Connectionism: {{Analysis}} of a Parallel Distributed Processing Model of Language Acquisition},
  shorttitle = {On Language and Connectionism},
  author = {Pinker, Steven and Prince, Alan},
  year = {1988},
  month = mar,
  journal = {Cognition},
  volume = {28},
  number = {1},
  pages = {73--193},
  issn = {0010-0277},
  doi = {10.1016/0010-0277(88)90032-7},
  urldate = {2023-03-23},
  langid = {english}
}

@misc{portelanceRolesNeuralNetworks2023,
  title = {The Roles of Neural Networks in Language Acquisitio},
  author = {Portelance, Eva and Jasbi, Masoud},
  year = {2023},
  doi = {10.31234/osf.io/b6978},
  urldate = {2023-12-19},
  copyright = {cc by}
}

@article{putnamMeaningMeaning1975,
  title = {The {{Meaning}} of '{{Meaning}}'},
  author = {Putnam, Hilary},
  year = {1975},
  journal = {Minnesota Studies in the Philosophy of Science},
  volume = {7},
  pages = {131--193},
  publisher = {{University of Minnesota Press}}
}

@inproceedings{qiuImprovingCompositionalGeneralization2022,
  title = {Improving {{Compositional Generalization}} with {{Latent Structure}} and {{Data Augmentation}}},
  booktitle = {Proceedings of the 2022 {{Conference}} of the {{North American Chapter}} of the {{Association}} for {{Computational Linguistics}}: {{Human Language Technologies}}},
  author = {Qiu, Linlu and Shaw, Peter and Pasupat, Panupong and Nowak, Pawel and Linzen, Tal and Sha, Fei and Toutanova, Kristina},
  year = {2022},
  month = jul,
  pages = {4341--4362},
  publisher = {{Association for Computational Linguistics}},
  address = {{Seattle, United States}},
  doi = {10.18653/v1/2022.naacl-main.323},
  urldate = {2023-08-21}
}

@article{quilty-dunnBestGameTown2022,
  title = {The {{Best Game}} in {{Town}}: {{The Re-Emergence}} of the {{Language}} of {{Thought Hypothesis Across}} the {{Cognitive Sciences}}},
  shorttitle = {The {{Best Game}} in {{Town}}},
  author = {{Quilty-Dunn}, Jake and Porot, Nicolas and Mandelbaum, Eric},
  year = {2022},
  month = dec,
  journal = {Behavioral and Brain Sciences},
  pages = {1--55},
  publisher = {{Cambridge University Press}},
  issn = {0140-525X, 1469-1825},
  doi = {10.1017/S0140525X22002849},
  urldate = {2023-03-21},
  langid = {english}
}

@article{raffelExploringLimitsTransfer2020,
  title = {Exploring the Limits of Transfer Learning with a Unified Text-to-Text Transformer},
  author = {Raffel, Colin and Shazeer, Noam and Roberts, Adam and Lee, Katherine and Narang, Sharan and Matena, Michael and Zhou, Yanqi and Li, Wei and Liu, Peter J.},
  year = {2020},
  month = jan,
  journal = {The Journal of Machine Learning Research},
  volume = {21},
  number = {1},
  pages = {140:5485--140:5551},
  issn = {1532-4435}
}

@misc{rameshHierarchicalTextConditionalImage2022,
  title = {Hierarchical {{Text-Conditional Image Generation}} with {{CLIP Latents}}},
  author = {Ramesh, Aditya and Dhariwal, Prafulla and Nichol, Alex and Chu, Casey and Chen, Mark},
  year = {2022},
  month = apr,
  number = {arXiv:2204.06125},
  eprint = {2204.06125},
  primaryclass = {cs},
  institution = {{arXiv}},
  doi = {10.48550/arXiv.2204.06125},
  urldate = {2022-06-02},
  archiveprefix = {arxiv}
}

@article{saltonVectorSpaceModel1975,
  title = {A Vector Space Model for Automatic Indexing},
  author = {Salton, G. and Wong, A. and Yang, C. S.},
  year = {1975},
  month = nov,
  journal = {Communications of the ACM},
  volume = {18},
  number = {11},
  pages = {613--620},
  issn = {0001-0782},
  doi = {10.1145/361219.361220},
  urldate = {2023-08-03}
}

@inproceedings{savelkaCanGPT4Support2023,
  title = {Can {{GPT-4 Support Analysis}} of {{Textual Data}} in {{Tasks Requiring Highly Specialized Domain Expertise}}?},
  booktitle = {Proceedings of the 2023 {{Conference}} on {{Innovation}} and {{Technology}} in {{Computer Science Education V}}. 1},
  author = {Savelka, Jaromir and Ashley, Kevin D. and Gray, Morgan A. and Westermann, Hannes and Xu, Huihui},
  year = {2023},
  month = jun,
  eprint = {2306.13906},
  primaryclass = {cs},
  pages = {117--123},
  doi = {10.1145/3587102.3588792},
  urldate = {2024-01-03},
  archiveprefix = {arxiv}
}

@inproceedings{savelkaThrilledYourProgress2023,
  title = {Thrilled by {{Your Progress}}! {{Large Language Models}} ({{GPT-4}}) {{No Longer Struggle}} to {{Pass Assessments}} in {{Higher Education Programming Courses}}},
  booktitle = {Proceedings of the 2023 {{ACM Conference}} on {{International Computing Education Research V}}.1},
  author = {Savelka, Jaromir and Agarwal, Arav and An, Marshall and Bogart, Chris and Sakr, Majd},
  year = {2023},
  month = aug,
  eprint = {2306.10073},
  primaryclass = {cs},
  pages = {78--92},
  doi = {10.1145/3568813.3600142},
  urldate = {2023-12-19},
  archiveprefix = {arxiv}
}

@book{schmidhuberCompositionalLearningDynamic1990,
  title = {Towards Compositional Learning with Dynamic Neural Networks},
  author = {Schmidhuber, J{\"u}rgen},
  year = {1990},
  publisher = {{Inst. f{\"u}r Informatik}}
}

@misc{schutBridgingHumanAIKnowledge2023,
  title = {Bridging the {{Human-AI Knowledge Gap}}: {{Concept Discovery}} and {{Transfer}} in {{AlphaZero}}},
  shorttitle = {Bridging the {{Human-AI Knowledge Gap}}},
  author = {Schut, Lisa and Tomasev, Nenad and McGrath, Tom and Hassabis, Demis and Paquet, Ulrich and Kim, Been},
  year = {2023},
  month = oct,
  number = {arXiv:2310.16410},
  eprint = {2310.16410},
  primaryclass = {cs, stat},
  publisher = {{arXiv}},
  doi = {10.48550/arXiv.2310.16410},
  urldate = {2023-12-19},
  archiveprefix = {arxiv}
}

@article{searleMindsBrainsPrograms1980,
  title = {Minds, {{Brains}}, and {{Programs}}},
  author = {Searle, John R.},
  year = {1980},
  journal = {Behavioral and Brain Sciences},
  volume = {3},
  number = {3},
  pages = {417--57},
  doi = {10.1017/s0140525x00005756}
}

@misc{shinnReflexionLanguageAgents2023,
  title = {Reflexion: {{Language Agents}} with {{Verbal Reinforcement Learning}}},
  shorttitle = {Reflexion},
  author = {Shinn, Noah and Cassano, Federico and Berman, Edward and Gopinath, Ashwin and Narasimhan, Karthik and Yao, Shunyu},
  year = {2023},
  month = oct,
  number = {arXiv:2303.11366},
  eprint = {2303.11366},
  primaryclass = {cs},
  publisher = {{arXiv}},
  doi = {10.48550/arXiv.2303.11366},
  urldate = {2023-10-23},
  archiveprefix = {arxiv}
}

@incollection{smolenskyConnectionismConstituentStructure1989,
  title = {Connectionism and {{Constituent Structure}}},
  booktitle = {Connectionism in {{Perspective}}},
  author = {Smolensky, Paul},
  editor = {Pfeifer, R. and Schreter, Z. and {Fogelman-Souli{\'e}}, F. and Steels, L.},
  year = {1989},
  month = aug,
  publisher = {{Elsevier}},
  googlebooks = {7pPv0STSos8C},
  isbn = {978-0-444-59876-9},
  langid = {english}
}

@article{smolenskyNeurocompositionalComputingCentral2022,
  title = {Neurocompositional {{Computing}}: {{From}} the {{Central Paradox}} of {{Cognition}} to a {{New Generation}} of {{AI Systems}}},
  shorttitle = {Neurocompositional {{Computing}}},
  author = {Smolensky, Paul and McCoy, Richard and Fernandez, Roland and Goldrick, Matthew and Gao, Jianfeng},
  year = {2022},
  month = sep,
  journal = {AI Magazine},
  volume = {43},
  number = {3},
  pages = {308--322},
  issn = {2371-9621},
  doi = {10.1002/aaai.12065},
  urldate = {2023-10-05},
  copyright = {Copyright (c) 2022 Paul Smolensky, Richard Thomas McCoy, Roland Fernandez, Matthew  Goldrick   , Jianfeng  Gao},
  langid = {english}
}

@article{smolenskyNeurocompositionalComputingHuman2022,
  title = {Neurocompositional Computing in Human and Machine Intelligence: {{A}} Tutorial},
  shorttitle = {Neurocompositional Computing in Human and Machine Intelligence},
  author = {Smolensky, Paul and McCoy, R. Thomas and Fernandez, Roland and Goldrick, Matthew and Gao, Jianfeng},
  year = {2022},
  publisher = {{Microsoft Technical Report MSR-TR-2022-5. https://www. microsoft. com/enus {\ldots}}},
  urldate = {2023-10-05}
}

@article{smolenskyProperTreatmentConnectionism1988,
  title = {On the Proper Treatment of Connectionism},
  author = {Smolensky, Paul},
  year = {1988},
  month = mar,
  journal = {Behavioral and Brain Sciences},
  volume = {11},
  number = {1},
  pages = {1--23},
  publisher = {{Cambridge University Press}},
  issn = {1469-1825, 0140-525X},
  doi = {10.1017/S0140525X00052432},
  urldate = {2023-08-21},
  langid = {english}
}

@incollection{soberMorganCanon1998,
  title = {Morgan's Canon},
  booktitle = {The Evolution of Mind},
  author = {Sober, Elliott},
  year = {1998},
  pages = {224--242},
  publisher = {{Oxford University Press}},
  address = {{New York, NY, US}},
  isbn = {978-0-19-511053-1}
}

@article{sullivanSAYCamLargeLongitudinal2021,
  title = {{{SAYCam}}: {{A Large}}, {{Longitudinal Audiovisual Dataset Recorded From}} the {{Infant}}'s {{Perspective}}},
  shorttitle = {{{SAYCam}}},
  author = {Sullivan, Jessica and Mei, Michelle and Perfors, Andrew and Wojcik, Erica and Frank, Michael C.},
  year = {2021},
  month = may,
  journal = {Open Mind},
  volume = {5},
  pages = {20--29},
  issn = {2470-2986},
  doi = {10.1162/opmi_a_00039},
  urldate = {2023-12-19}
}

@article{thirunavukarasuLargeLanguageModels2023,
  title = {Large Language Models in Medicine},
  author = {Thirunavukarasu, Arun James and Ting, Darren Shu Jeng and Elangovan, Kabilan and Gutierrez, Laura and Tan, Ting Fang and Ting, Daniel Shu Wei},
  year = {2023},
  month = aug,
  journal = {Nature Medicine},
  volume = {29},
  number = {8},
  pages = {1930--1940},
  publisher = {{Nature Publishing Group}},
  issn = {1546-170X},
  doi = {10.1038/s41591-023-02448-8},
  urldate = {2024-01-03},
  copyright = {2023 Springer Nature America, Inc.},
  langid = {english}
}

@book{tomaselloConstructingLanguage2009,
  title = {Constructing a {{Language}}},
  author = {Tomasello, Michael},
  year = {2009},
  month = jun,
  publisher = {{Harvard University Press}},
  googlebooks = {7M\_lSEfzTQoC},
  isbn = {978-0-674-04439-5},
  langid = {english}
}

@misc{touvronLlamaOpenFoundation2023,
  title = {Llama 2: {{Open Foundation}} and {{Fine-Tuned Chat Models}}},
  shorttitle = {Llama 2},
  author = {Touvron, Hugo and Martin, Louis and Stone, Kevin and Albert, Peter and Almahairi, Amjad and Babaei, Yasmine and Bashlykov, Nikolay and Batra, Soumya and Bhargava, Prajjwal and Bhosale, Shruti and Bikel, Dan and Blecher, Lukas and Ferrer, Cristian Canton and Chen, Moya and Cucurull, Guillem and Esiobu, David and Fernandes, Jude and Fu, Jeremy and Fu, Wenyin and Fuller, Brian and Gao, Cynthia and Goswami, Vedanuj and Goyal, Naman and Hartshorn, Anthony and Hosseini, Saghar and Hou, Rui and Inan, Hakan and Kardas, Marcin and Kerkez, Viktor and Khabsa, Madian and Kloumann, Isabel and Korenev, Artem and Koura, Punit Singh and Lachaux, Marie-Anne and Lavril, Thibaut and Lee, Jenya and Liskovich, Diana and Lu, Yinghai and Mao, Yuning and Martinet, Xavier and Mihaylov, Todor and Mishra, Pushkar and Molybog, Igor and Nie, Yixin and Poulton, Andrew and Reizenstein, Jeremy and Rungta, Rashi and Saladi, Kalyan and Schelten, Alan and Silva, Ruan and Smith, Eric Michael and Subramanian, Ranjan and Tan, Xiaoqing Ellen and Tang, Binh and Taylor, Ross and Williams, Adina and Kuan, Jian Xiang and Xu, Puxin and Yan, Zheng and Zarov, Iliyan and Zhang, Yuchen and Fan, Angela and Kambadur, Melanie and Narang, Sharan and Rodriguez, Aurelien and Stojnic, Robert and Edunov, Sergey and Scialom, Thomas},
  year = {2023},
  month = jul,
  number = {arXiv:2307.09288},
  eprint = {2307.09288},
  primaryclass = {cs},
  publisher = {{arXiv}},
  doi = {10.48550/arXiv.2307.09288},
  urldate = {2023-09-15},
  archiveprefix = {arxiv}
}

@article{tshitoyanUnsupervisedWordEmbeddings2019,
  title = {Unsupervised Word Embeddings Capture Latent Knowledge from Materials Science Literature},
  author = {Tshitoyan, Vahe and Dagdelen, John and Weston, Leigh and Dunn, Alexander and Rong, Ziqin and Kononova, Olga and Persson, Kristin A. and Ceder, Gerbrand and Jain, Anubhav},
  year = {2019},
  month = jul,
  journal = {Nature},
  volume = {571},
  number = {7763},
  pages = {95--98},
  publisher = {{Nature Publishing Group}},
  issn = {1476-4687},
  doi = {10.1038/s41586-019-1335-8},
  urldate = {2023-12-19},
  copyright = {2019 The Author(s), under exclusive licence to Springer Nature Limited},
  langid = {english}
}

@article{turingComputingMachineryIntelligence1950,
  title = {Computing {{Machinery}} and {{Intelligence}}},
  author = {Turing, A. M.},
  year = {1950},
  journal = {Mind},
  volume = {59},
  number = {236},
  eprint = {2251299},
  eprinttype = {jstor},
  pages = {433--460},
  publisher = {{[Oxford University Press, Mind Association]}},
  issn = {0026-4423},
  urldate = {2020-10-30}
}

@incollection{vaswaniAttentionAllYou2017,
  title = {Attention Is {{All}} You {{Need}}},
  booktitle = {Advances in {{Neural Information Processing Systems}} 30},
  author = {Vaswani, Ashish and Shazeer, Noam and Parmar, Niki and Uszkoreit, Jakob and Jones, Llion and Gomez, Aidan N and Kaiser, {\L}ukasz and Polosukhin, Illia},
  editor = {Guyon, I. and Luxburg, U. V. and Bengio, S. and Wallach, H. and Fergus, R. and Vishwanathan, S. and Garnett, R.},
  year = {2017},
  pages = {5998--6008},
  publisher = {{Curran Associates, Inc.}},
  urldate = {2020-09-11}
}

@inproceedings{wallaceNLPModelsKnow2019,
  title = {Do {{NLP Models Know Numbers}}? {{Probing Numeracy}} in {{Embeddings}}},
  shorttitle = {Do {{NLP Models Know Numbers}}?},
  booktitle = {Proceedings of the 2019 {{Conference}} on {{Empirical Methods}} in {{Natural Language Processing}} and the 9th {{International Joint Conference}} on {{Natural Language Processing}} ({{EMNLP-IJCNLP}})},
  author = {Wallace, Eric and Wang, Yizhong and Li, Sujian and Singh, Sameer and Gardner, Matt},
  editor = {Inui, Kentaro and Jiang, Jing and Ng, Vincent and Wan, Xiaojun},
  year = {2019},
  month = nov,
  pages = {5307--5315},
  publisher = {{Association for Computational Linguistics}},
  address = {{Hong Kong, China}},
  doi = {10.18653/v1/D19-1534},
  urldate = {2023-12-14}
}

@misc{wangByteSized32CorpusChallenge2023,
  title = {{{ByteSized32}}: {{A Corpus}} and {{Challenge Task}} for {{Generating Task-Specific World Models Expressed}} as {{Text Games}}},
  shorttitle = {{{ByteSized32}}},
  author = {Wang, Ruoyao and Todd, Graham and Yuan, Eric and Xiao, Ziang and C{\^o}t{\'e}, Marc-Alexandre and Jansen, Peter},
  year = {2023},
  month = oct,
  number = {arXiv:2305.14879},
  eprint = {2305.14879},
  primaryclass = {cs},
  publisher = {{arXiv}},
  doi = {10.48550/arXiv.2305.14879},
  urldate = {2023-12-11},
  archiveprefix = {arxiv}
}

@misc{wangDocumentLevelMachineTranslation2023a,
  title = {Document-{{Level Machine Translation}} with {{Large Language Models}}},
  author = {Wang, Longyue and Lyu, Chenyang and Ji, Tianbo and Zhang, Zhirui and Yu, Dian and Shi, Shuming and Tu, Zhaopeng},
  year = {2023},
  month = apr,
  number = {2304.02210},
  eprint = {2304.02210},
  publisher = {{arXiv}},
  urldate = {2024-01-03},
  archiveprefix = {arxiv},
  langid = {english}
}

@inproceedings{warstadtFindingsBabyLMChallenge2023,
  title = {Findings of the {{BabyLM Challenge}}: {{Sample-Efficient Pretraining}} on {{Developmentally Plausible Corpora}}},
  shorttitle = {Findings of the {{BabyLM Challenge}}},
  booktitle = {Proceedings of the {{BabyLM Challenge}} at the 27th {{Conference}} on {{Computational Natural Language Learning}}},
  author = {Warstadt, Alex and Mueller, Aaron and Choshen, Leshem and Wilcox, Ethan and Zhuang, Chengxu and Ciro, Juan and Mosquera, Rafael and Paranjabe, Bhargavi and Williams, Adina and Linzen, Tal and Cotterell, Ryan},
  editor = {Warstadt, Alex and Mueller, Aaron and Choshen, Leshem and Wilcox, Ethan and Zhuang, Chengxu and Ciro, Juan and Mosquera, Rafael and Paranjabe, Bhargavi and Williams, Adina and Linzen, Tal and Cotterell, Ryan},
  year = {2023},
  month = dec,
  pages = {1--6},
  publisher = {{Association for Computational Linguistics}},
  address = {{Singapore}},
  urldate = {2023-12-09}
}

@incollection{warstadtWhatArtificialNeural2022,
  title = {What {{Artificial Neural Networks Can Tell Us}} about {{Human Language Acquisition}}},
  booktitle = {Algebraic {{Structures}} in {{Natural Language}}},
  author = {Warstadt, Alex and Bowman, Samuel R.},
  year = {2022},
  publisher = {{CRC Press}},
  isbn = {978-1-00-320538-8}
}

@incollection{weaverTranslation1955,
  title = {Translation},
  booktitle = {Machine {{Translation}} of {{Languages}}},
  author = {Weaver, Warren},
  editor = {Locke, William N. and Booth, Donald A.},
  year = {1955},
  month = may,
  publisher = {{MIT Press}},
  address = {{Boston, MA}},
  isbn = {978-0-262-12002-9}
}

@article{weiChainofThoughtPromptingElicits2022,
  title = {Chain-of-{{Thought Prompting Elicits Reasoning}} in {{Large Language Models}}},
  author = {Wei, Jason and Wang, Xuezhi and Schuurmans, Dale and Bosma, Maarten and Ichter, Brian and Xia, Fei and Chi, Ed and Le, Quoc V. and Zhou, Denny},
  year = {2022},
  month = dec,
  journal = {Advances in Neural Information Processing Systems},
  volume = {35},
  pages = {24824--24837},
  urldate = {2023-09-28},
  langid = {english}
}

@article{winogradProceduresRepresentationData1971,
  title = {Procedures as a {{Representation}} for {{Data}} in a {{Computer Program}} for {{Understanding Natural Language}}},
  author = {Winograd, Terry},
  year = {1971},
  month = jan,
  urldate = {2023-03-28},
  langid = {american}
}

@book{wittgensteinPhilosophicalInvestigations1953,
  title = {Philosophical {{Investigations}}},
  author = {Wittgenstein, Ludwig},
  year = {1953},
  publisher = {{Wiley-Blackwell}},
  address = {{New York, NY, USA}}
}

@misc{zengSocraticModelsComposing2022,
  title = {Socratic {{Models}}: {{Composing Zero-Shot Multimodal Reasoning}} with {{Language}}},
  shorttitle = {Socratic {{Models}}},
  author = {Zeng, Andy and Attarian, Maria and Ichter, Brian and Choromanski, Krzysztof and Wong, Adrian and Welker, Stefan and Tombari, Federico and Purohit, Aveek and Ryoo, Michael and Sindhwani, Vikas and Lee, Johnny and Vanhoucke, Vincent and Florence, Pete},
  year = {2022},
  month = may,
  number = {arXiv:2204.00598},
  eprint = {2204.00598},
  primaryclass = {cs},
  publisher = {{arXiv}},
  doi = {10.48550/arXiv.2204.00598},
  urldate = {2023-12-19},
  archiveprefix = {arxiv}
}

@misc{zhangBenchmarkingLargeLanguage2023a,
  title = {Benchmarking {{Large Language Models}} for {{News Summarization}}},
  author = {Zhang, Tianyi and Ladhak, Faisal and Durmus, Esin and Liang, Percy and McKeown, Kathleen and Hashimoto, Tatsunori B.},
  year = {2023},
  month = jan,
  number = {arXiv:2301.13848},
  eprint = {2301.13848},
  publisher = {{arXiv}},
  urldate = {2024-01-03},
  archiveprefix = {arxiv},
  langid = {english}
}

@article{zhangUnderstandingDeepLearning2021,
  title = {Understanding Deep Learning (Still) Requires Rethinking Generalization},
  author = {Zhang, Chiyuan and Bengio, Samy and Hardt, Moritz and Recht, Benjamin and Vinyals, Oriol},
  year = {2021},
  month = feb,
  journal = {Communications of the ACM},
  volume = {64},
  number = {3},
  pages = {107--115},
  issn = {0001-0782},
  doi = {10.1145/3446776},
  urldate = {2023-12-19}
}

@misc{zhouSolvingChallengingMath2023a,
  title = {Solving {{Challenging Math Word Problems Using GPT-4 Code Interpreter}} with {{Code-based Self-Verification}}},
  author = {Zhou, Aojun and Wang, Ke and Lu, Zimu and Shi, Weikang and Luo, Sichun and Qin, Zipeng and Lu, Shaoqing and Jia, Anya and Song, Linqi and Zhan, Mingjie and Li, Hongsheng},
  year = {2023},
  month = aug,
  number = {arXiv:2308.07921},
  eprint = {2308.07921},
  primaryclass = {cs},
  publisher = {{arXiv}},
  doi = {10.48550/arXiv.2308.07921},
  urldate = {2023-12-19},
  archiveprefix = {arxiv}
}

\end{document}